\pdfoutput=1

\documentclass[11pt]{article}

\usepackage[]{ACL2023}

\usepackage{times}
\usepackage{latexsym}

\usepackage[T1]{fontenc}

\usepackage[utf8]{inputenc}

\usepackage{microtype}

\usepackage{inconsolata}

\usepackage{bm}
\usepackage{multirow}
\usepackage{subfig}
\usepackage{algorithmic}
\usepackage{amsfonts}
\usepackage{booktabs}
\usepackage{graphicx}
\usepackage{comment}
\usepackage{hyperref}
\usepackage{fancyvrb}
\usepackage{cancel}

\title{Leveraging Codebook Knowledge with NLI and ChatGPT for Zero-Shot Political Relation Classification}

\author{Yibo Hu$^*$, Erick Skorupa Parolin$^\dagger$, Latifur Khan$^\dagger$, Patrick T.~Brandt$^\dagger$,\\ 
\textbf{Javier Osorio$^\ddagger$,  Vito J.~D'Orazio$^\mathsection$}\\
    $^*$Georgia Institute of Technology, $^\dagger$The University of Texas at Dallas,\\ $^\ddagger$The University of Arizona,
    $^\mathsection$West Virginia University\\
    \texttt{yibo.hu@gatech.edu}, \texttt{erickparolin@gmail.com},
    \texttt{\{lkhan,pbrandt\}@utdallas.edu}, \\
    \texttt{josorio1@email.arizona.edu, vito.dorazio@mail.wvu.edu}
}

\begin{document}
\maketitle

\begin{abstract}

Is it possible accurately classify political relations within evolving event ontologies without extensive annotations?  
This study investigates zero-shot learning methods that use expert knowledge from existing annotation codebook, and evaluates the performance of advanced ChatGPT (GPT-3.5/4) and a natural language inference (NLI)-based model called ZSP.  
ChatGPT uses codebook's labeled summaries as prompts, whereas ZSP breaks down the classification task into context, event mode, and class disambiguation to refine task-specific hypotheses. This decomposition enhances interpretability, efficiency, and adaptability to schema changes.
The experiments reveal ChatGPT's strengths and limitations, and crucially show ZSP's outperformance of dictionary-based methods and its competitive edge over some supervised models. 
These findings affirm the value of ZSP for validating event records and advancing ontology development. Our study underscores the efficacy of leveraging transfer learning and existing domain expertise to enhance research efficiency and scalability.  
The code is publicly available\footnote{{\url{https://github.com/snowood1/Zero-Shot-PLOVER}}}.
\end{abstract}

\section{Introduction}

Event coding is a crucial task in political violence research for both academic and policy communities. It transforms unstructured text from news articles into structured event data, represented as source-action-target triplets, achieved  through entity extraction and relation classification. 
It provides a structured record of interactions among political actors and serves as input for monitoring, understanding, and forecasting political conflicts and mediation processes worldwide \cite{schrodt1996using,schrodt2003evaluating,schrodt2004using,schrodt1997early,schrodt2006forecasting,schrodt2011forecasting,shellman2007predicting,shearer2007forecasting,brandt2011real,brandt2013forecasting,brandt2014evaluating}. 
However, manually coding events from extensive datasets is labor-intensive.

To streamline this, experts have developed event ontologies and knowledge bases \cite{mcclelland1978world,azar1980conflict,cameo,IDEA,Schrodt2006,Boschee2016,UPetrarch,osorio_enhancing_2020,osorio_mapping_2019}.  	
Yet, traditional pattern-matching models based on static dictionaries suffer from inflexibility, low recall, and high maintenance costs. 
Recent advancements in deep learning and pretrained language models (PLMs) offer promising supervised learning solutions \cite{glavas-etal-2017-cross, buyukoz-etal-2020-analyzing, olsson-etal-2020-text,ors-etal-2020-event,parolin2019hanke, parolin20213m,parolin2022ConfliT5,conflibert2022}. 
Yet, their reliance on extensively annotated datasets introduces significant challenges, especially for in-depth and subnational studies requiring nuanced categorization and non-exclusive labeling within political event ontologies. 
Moreover, labeled datasets lack flexibility and may require frequent relabeling as ontologies evolve. 
Thus, much of the current PLM-based research in event coding targets broad, coarse-grained categorizations, often constrained by limited evaluation sets.
Costs, time, and effort associated with developing training data have foiled the large-scale adoption and ready deployment of PLMs by government security agencies, researchers, and practitioners in need of monitoring and understanding rapidly-changing conflict processes around the world.

In light of these challenges, we pose the following questions:
(1) Is it possible to leverage existing expert knowledge to enhance the efficiency of event coding without extensive annotation of new data?
(2) Is it possible to create an interpretable and adaptable system that easily accommodates ontology or schema changes?

To tackle these questions, our paper focuses on relation classification, a key aspect of event coding. 
The goal is to categorize events in a source-target pair following a predefined event ontology PLOVER \cite{plover} without external labeled data. 
We achieved this by combining the transferred semantic knowledge of PLMs with expertise derived from annotation codebooks.
The codebook, as depicted in Figure \ref{fig:framework}, contains label descriptions and guidelines  for resolving  confusing labels. 
To unlock this knowledge,  we explore two zero-shot methods: the emerging ChatGPT (GPT-3.5/4),  and our proposed NLI-based model called ZSP (\textbf{Z}ero-\textbf{S}hot fine-grained relation classification model for \textbf{P}LOVER ontology). 

While GPT-4 showcases notable improvements over GPT-3.5, it still exhibits instability in fine-grained tasks, promising further enhancement. 
Conversely, ZSP, despite being built upon a smaller model, offers substantial advantages. It leverages easily constructed hypotheses from the codebook and employs a tree-query framework to capture nuanced semantics and mode distinctions within a focused set of hypotheses at each level. Additionally, ZSP's adaptability allows straightforward updates by modifying the hypothesis table or class disambiguation rules to align with evolving ontologies. This approach proves more cost-effective than maintaining extensive dictionaries or re-labeling datasets for event record validation.

\begin{table}\setlength{\tabcolsep}{2pt}
    \centering
\resizebox{\columnwidth}{!}{
\begin{tabular}{lcc} 
\toprule
\multicolumn{1}{c}{CAMEO Root.}    & PLOVER Root.    & Quad.    \\ 
\midrule
01- Make Public Statement    & dropped    &    \\
02- Appeal    & dropped    &    \\
03- Express Intent to Cooperate    & AGREE    & 1. V-Coop.    \\
04- Consult    & CONSULT    & 1. V-Coop.    \\
05- Engage in Diplomatic Cooperation & SUPPORT    & 1. V-Coop.    \\
06- Engage in Material Cooperation   & COOPERATE    & 2. M-Coop.   \\
07- Provide Aid    & AID    & 2. M-Coop.   \\
08- Yield           & YIELD    & 2. M-Coop.   \\
09- Investigate    & ACCUSE    & 3. V-Conf.   \\
10- Demand    & REQUEST    & 3. V-Conf.   \\
11- Disapprove    & ACCUSE    & 3. V-Conf.   \\
12- Reject    & REJECT    & 3. V-Conf.   \\
13- Threaten    & THREATEN    & 3. V-Conf.   \\
14- Protest    & PROTEST    & 4. M-Conf.   \\
15- Exhibit Force Posture    & MOBILIZE    & 4. M-Conf.   \\
16- Reduce Relations    & SANCTION    & 4. M-Conf.   \\
17- Coerce    & COERCE    & 4. M-Conf.   \\
18- Assault    & ASSAULT & 4. M-Conf.  \\
20- Unconventional Mass Violence    & ASSAULT & 4. M-Conf.  \\
\bottomrule
\end{tabular}
}
    \caption{CAMEO/PLOVER's Rootcodes and Quadcodes (1-Verbal Cooperation, 2-Material Cooperation, 3-Verbal Conflict, and 4-Material Conflict).}
    \label{tab:preliminaries}
\end{table}

In sum, the untapped potential of GPT-4 and the success of ZSP encourage experts to reevaluate the value of existing knowledge bases and inspire innovative uses of this knowledge to expedite research within the political science community.

\section{Preliminaries}
\label{sec:Preliminaries}

\subsection{Event Coding, Ontology, and Mode}\label{sec:event ontology}

The ontology of the event coding system defines how to code the actors  \cite{mcclelland1978world,azar1980conflict,MID,IDEA,acled,mitamura2015tac,Boschee2016}.
One prominent schema is \textbf{CAMEO} \cite{cameo}, which incorporates knowledge from the codebook\footnote{{\url{https://eventdata.parusanalytics.com/cameo.dir/ CAMEO.Manual.1.1b3.pdf}}}, action-pattern dictionaries, and actor dictionaries. It categorizes political interactions into 200+ fine-grained 4-digit codes (\texttt{01XX}--\texttt{20XX}). These are then aggregated into 20 more frequently utilized \textbf{Rootcodes}  (\texttt{01}--\texttt{20}), and further into 4 high-level \textbf{Quadcodes}: 1-Verbal Cooperation, 2-Material Cooperation, 3-Verbal Conflict, and 4-Material Conflict. 
Later, the \textbf{PLOVER} scheme \cite{plover} simplifies CAMEO by removing 4-digit codes, reducing Rootcodes to 16, and enhancing semantic clarity. 
Rootcode and Quadcode overviews of CAMEO/PLOVER are presented in Table \ref{tab:preliminaries}, with a codebook snippet in Figure \ref{fig:framework}. 
Appendix Table \ref{tab:chatgpt} and the codebook present the Rootcode in detail.

Figure \ref{fig:cameo example} illustrates an event coding scenario where the interaction between the source (Obama representing the USA government) and the target (Israel, coded as ISR) is classified using lower-level 4-digit codes as well as higher-level Rootcodes and Quadcodes.  However, the classification is highly sensitive to subtle variations in what we term \textbf{event mode}—whether Obama has provided, plans to provide, has stopped, or intends to stop military aid—resulting in significant adjustments to Rootcodes and Quadcodes for identical entities.

\begin{figure}\fontsize{7.3pt}{8pt}
\centering
\begin{BVerbatim}[commandchars=\\\{\}]
\textcolor{red}{Obama} said he won't provide military aid to \textcolor{blue}{Israel}.
\textbf{Source:}    \textcolor{red}{Obama-USAGOV}                \textbf{Target:}   \textcolor{blue}{Israel-ISR}
\textbf{Action:}    1222-Reject request for military aid 
\textbf{Root.}:     12- REJECT                    \textbf{Quad.}:    3. V-Conf.

\textbf{Other event modes} -------------------- \textbf{Root.} ----- \textbf{Quad.} -
\textcolor{red}{Obama} halted   military aid to \textcolor{blue}{Israel}. SANCTION  4. M-Conf.
\textcolor{red}{Obama} provided military aid to \textcolor{blue}{Israel}. AID       2. M-Coop.
\textcolor{red}{Obama} agreed to provide aid to \textcolor{blue}{Israel}. AGREE     1. V-Coop.
\end{BVerbatim}
\caption{Event coding illustration: How event modes affect Rootcode and Quadcode labeling for sentences involving identical entities.}\label{fig:cameo example}
\end{figure}

PLOVER's codebook\footnote{{\url{https://github.com/openeventdata/PLOVER/blob/master/PLOVER_MANUAL.pdf}}} suggested  the concept of event mode through auxiliary modes—historical, future, hypothetical, or negated—to flexibly depict event statuses and assist in labeling. However, it did so without providing strict definitions or implementations. Inspired by this, we refine these event modes and integrate them into our NLI system to enhance classification accuracy. This refinement process and its impact on enhancing event coding precision are detailed in Section \ref{sec:3.2} and further elaborated in Appendix \ref{appendix:PLOVER2mode_mapping}.

The shift from the dictionary-based CAMEO to the more semantically friendly PLOVER aligns with the domain's broader trend. Our focus on PLOVER is the result of careful consideration and validation with domain experts.

\subsection{Related Work}\label{s:related work}

Relation or event extraction has been studied across various domains \cite{hendrickx2019semeval,zhang2017position,han-etal-2018-fewrel,MUC_4_cleaned,luan2018multi,nyt_dataset,fincke2022language},  with some studies partially overlapping in entities or categorizations relevant to political science \cite{ace04,RAMS,wikievent}. 
However, our work distinguishes itself by considering event modes, a dimension not fully explored in existing works.

Our work also relates to zero-shot learning across various schemes \cite{huang2018zero,obamuyide-vlachos-2018-zero,yin2019benchmarking,meng2020text,geng2021ontozsl,lyu2021zero,sainz2021label}, especially socio-political event classification \cite{case2021,radford-2021-case,barker-etal-2021-ibm,haneczok2021fine,halterman2021few}. 
However, many works focus on  sentence-level classification rather than relations between multiple entity pairs. 
The others with complex templates \textbf{cannot be} adapted to our political ontology easily. Thus, we design our framework to efficiently integrate with the existing knowledge base.

Finally, recent large language models (LLMs) \cite{brown2020language,instructGPT,chatgpt,gpt4} have greatly advanced zero-shot learning in reasoning and text generation \cite{hu2022controllable,halim2023wokegpt,jin2024better}. However, the application of ChatGPT for zero-shot event extraction remains underexplored and lags behind advanced supervised methods \cite{yuan2023zero, Cai2023monte, Li2023evaluating, Gao2023exploring,aiyappa2023can}. We will evaluate ChatGPT on PLOVER as part of our investigation.

\section{Approach}

We start by discussing the discovery of NLI as a potential solution and the construction process of the NLI-based ZSP framework, followed by the deployment of ChatGPT.

\subsection{Limitations of NLI for Event Coding}\label{sec:nli}

NLI measures how likely a premise entails a hypothesis \cite{nli,multinli}. 
Initially, we explores the feasibility of using NLI to assign PLOVER codes by selecting the most probable entailed hypothesis from a set of candidates. 
We designed a tiny experiment with only 18 hypotheses derived from the Rootcode names\footnote{See more details about this ``Tiny model'' experiment in Section \ref{sec:ablation study}, and Rootcode names in Table \ref{tab:preliminaries}).}. 

Table \ref{tab:nli examples} illustrates three example hypotheses,  where \textless{}S\textgreater{} and \textless{}T\textgreater{} denote the source (Indonesian students) and the target (President Suharto’s government), respectively. The labeled premise ``THREATEN 3''  indicates the intention to initiate protests.  Notably, NLI accurately recognizes AID as contradictory and identifies REQUEST and PROTEST as entailments.  Moreover, the tiny NLI model with only 18 hypotheses  surpasses dictionary-based methods that rely on $81k$ verb patterns, with a remarkable macro F1 increase of 17.1\% for Quadcode classification, thus comfirming the NLI potential as a valuable solution.

However, upon closer examination, we find that NLI models measuring semantic entailment may not directly suit our classification task, as the best entailed hypotheses do not always match our desired labels. The adaptation raises two key issues:

First, NLI disregards event mode.
In Table \ref{tab:nli examples}, the premise labeled as THREATEN stands for a hypothetical, verbal protest event. NLI  partially captures the event's context (PROTEST) but fails to consider its mode. 
To address this, we incorporate mode information to enhance candidate precision.

\begin{figure*}[t]
    \centering
    \includegraphics[width=\linewidth]{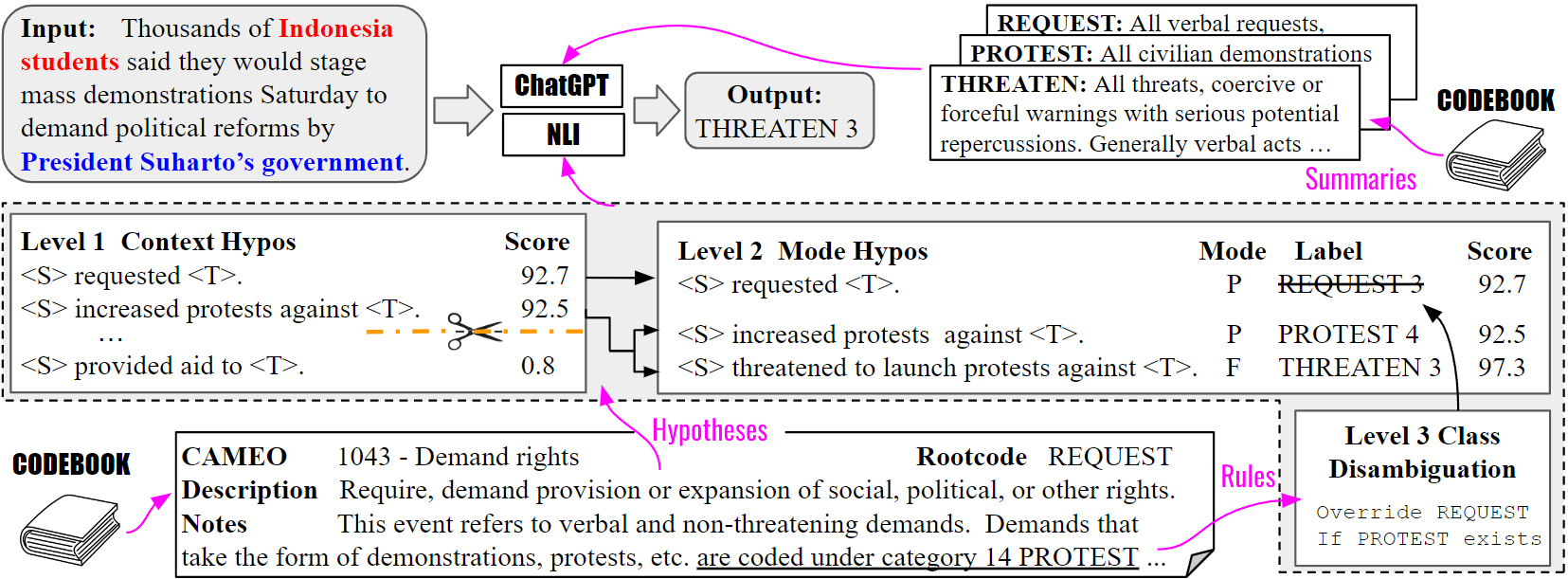}
    \caption{Two zero-shot approaches for classifying relation labels (Rootcode and Quadcode) in a \textcolor{red}{source} - \textcolor{blue} {target} pair. ChatGPT employs prompts designed from the codebook's label summaries, while ZSP utilizes a pretrained NLI model and a tree-query system. Hypotheses and class disambiguation rules are derived from the codebook and enhanced with mode considerations (e.g., \textbf{P}ast, \textbf{F}uture). The tree-query framework reduces query time and improves precision by filtering candidates, determining modes, and eliminating ambiguity.
    }
    \label{fig:framework}
\end{figure*}

Second, event category labels lack mutual exclusivity  in semantics.
In Table \ref{tab:nli examples}, the premise correctly entails both PROTEST and REQUEST with high scores from the semantic aspect. However, in CAMEO/PLOVER's single-label schema, the context ``demonstrate to demand reforms'' aligns with PROTEST, a Material Conflict,  rather than REQUEST, a Verbal Conflict. An easy solution is to prioritize ``protest'' over ``request''  when encountering ``protest to request'', following the codebook's disambiguation rules illustrated in Figure \ref{fig:framework}.

In summary, we identify three key dimensions to ensure accurate predictions: Context, Mode, and Class Disambiguation. 
Firstly, we narrow down predictions to the top candidates, PROTEST and REQUEST. Secondly, we incorporate mode information and identify the event as future, verbal, or hypothetical. Lastly, we apply the class disambiguation rule, giving precedence to PROTEST over REQUEST. By combining these dimensions, we achieve the final correct answer THREATEN. These findings motivate our NLI-based framework in Figure \ref{fig:framework}. Next, we provide detailed explanations for each component.

\begin{table}[t]\setlength{\tabcolsep}{1pt}
\centering
\resizebox{\columnwidth}{!}{
\begin{tabular}{llrr} 
\toprule
\multicolumn{4}{p{1.3\columnwidth}}{\textbf{Premise: } Thousands of \textcolor{red}{Indonesian students} said they would stage mass demonstrations Saturday, demanding political reforms from \textcolor{blue}{President Suharto’s government}.} \\
\multicolumn{4}{p{1.3\columnwidth}}{\textbf{Source <S>:} Indonesian students} \\ 
\multicolumn{4}{p{1.3\columnwidth}}{\textbf{Target <T>:} President Suharto’s government}\\ 
\multicolumn{4}{p{1.3\columnwidth}}{\textbf{Gold Label:} THREATEN 3; threaten political dissent.} \\ 

\midrule
\textbf{(a) Basic Hypotheses} & {\qquad} & \multicolumn{1}{c}{\textbf{Label}} & \textbf{Score}   \\ 
\midrule
\textless{}S\textgreater{} requested \textless{}T\textgreater{}.    &  & REQUEST 3    & 92.7    \\
\textless{}S\textgreater{} protested against \textless{}T\textgreater{}.    &  & PROTEST 4    & 92.5    \\
\textless{}S\textgreater{} provided aid to \textless{}T\textgreater{}.    &  & AID 2    & 0.8    \\
\midrule
\multicolumn{4}{l}{\textbf{(b) Adding Mode}}  \\ 
\textless{}S\textgreater{} threatened to protest against \textless{}T\textgreater{}.    &  & \textbf{\checkmark  THREAT. 3}    & \textbf{97.3}   \\
\midrule
\multicolumn{4}{l}{\textbf{(c) Adding Class Disambiguation}}  \\ 
\multicolumn{4}{l}{\texttt{Override REQUEST if PROTEST exists} $\rightarrow$ \cancel{REQUEST}} \\ 

\bottomrule
\end{tabular}
}

\caption{Entailment scores (\%) for hypotheses on a sentence labeled as ``THREATEN 3'' (Rootcode text + Quadcode digit). Adding mode or class disambiguation to basic hypotheses improves prediction precision.}
\label{tab:nli examples}

\end{table}

\subsection{Enabling NLI to Classify Event Mode}\label{sec:3.2}

NLI's inability to accurately determine the event mode often leads to misclassification. In Table \ref{tab:nli mode examples}, we present  a sentence with reversed labels compared to another in Table \ref{tab:nli examples}. Although labeled as AGREE for Verbal Cooperation, indicating a willingness to mitigate dissent, it incorrectly scores high (76.4\%) for the hypothesis ``protested against''. Semantically, this isn't entirely wrong, as ``agreed to ease protests'' implies prior protests, but it suggests cooperation rather than the implied conflict of the hypothesis.

To address this, we introduce \textbf{mode-aware hypotheses} that incorporate four event modess, adapted from PLOVER's guidelines:  Past (\textbf{P}) for historical events or events that initiated or are ongoing, Future (\textbf{F}) for future, verbal, or hypothetical events, Contradict\_Past (\textbf{CP}) and Contradict\_Future (\textbf{CF}) for their respective contradictions. See examples in Table \ref{tab:nli mode examples}.

Labels for each mode are directly adopted from the codebook, requiring no new definitions. 
For example, PROTEST's CF is labeled as AGREE, mirroring ``03- EXPRESS INTENT TO COOPERATE'' Rootcode, specifically item ``0352- Express intent to ease popular dissent.'' 
Likewise, the CP events like ``reduced protests against'' are labeled as YIELD, following CAMEO code 0833, which signifies yielding to demands. 

To facilitate this nuanced classification approach, we have assembled a mode mapping table (Table \ref{tab:plover_mode}) that clarifies mode transitions and associated label modifications. This streamlined process not only simplifies the task of classification but also significantly reduces the complexity of navigating the codebook. For an in-depth exploration of these event modes, please see Appendix   \ref{appendix:PLOVER2mode_mapping}.

\begin{table}[t]\setlength{\tabcolsep}{1pt}
\centering
\resizebox{\columnwidth}{!}{
\begin{tabular}{lcrr} 
\toprule
\multicolumn{4}{p{1.4\columnwidth}}{\textbf{Premise:} Thousands of \textcolor{red}{Indonesian students} agreed to suspend Saturday's demonstrations, demanding political reforms from \textcolor{blue}{President Suharto's government}.} \\ 

\multicolumn{4}{p{1.3\columnwidth}}{\textbf{Source <S>:} Indonesian students} \\ 
\multicolumn{4}{p{1.3\columnwidth}}{\textbf{Target <T>:} President Suharto’s government}\\ 

\multicolumn{4}{p{1.4\columnwidth}}{\textbf{Gold Label:} AGREE 1; express intent to ease popular dissent.} \\ 

\midrule

\begin{tabular}[c]{@{}l@{}}\textbf{Mode Hypotheses for ``Protest''}\end{tabular} & \textbf{Mode} & \multicolumn{1}{c}{\textbf{Label}} & \multicolumn{1}{c}{\textbf{Score}}    \\ 
\midrule
\textless{}S\textgreater{} protested against \textless{}T\textgreater{}.    & -    & PROTEST 4    & 92.5    \\
\midrule

\textless{}S\textgreater{} increased protests against \textless{}T\textgreater{}.    & {P}    & {PROTEST 4}    & 0.1    \\
\textless{}S\textgreater{} launched more protests against \textless{}T\textgreater{}.    & P &  {PROTEST 4}     & 0.0    \\

\textless{}S\textgreater{} reduced protests against \textless{}T\textgreater{}.    & CP    &  YIELD 2    & 95.2 \\

\textless{}S\textgreater{} threatened to protest against  \textless{}T\textgreater{}.    & F   & THREAT. 3    & 67.5   \\

\textless{}S\textgreater{} promised to reduce protests against \textless{}T\textgreater{}.    & \textbf{CF} & \textbf{AGREE 1}  & \textbf{97.1}    \\
\textless{}S\textgreater{} will reduce protests against \textless{}T\textgreater{}. & CF  &   AGREE 1  & 96.3   \\  

\bottomrule
\end{tabular}
}
\caption{Entailment scores (\%) for hypotheses on a sentence labeled as ``AGREE 1'' (Rootcode text + Quadcode digit). Adding \textbf{Mode} (P, F, CP, CF) improves prediction precision compared to mode exclusion (-).} 
\label{tab:nli mode examples}
\end{table}

The mode-aware hypotheses in Table \ref{tab:nli mode examples} enable NLI to accurately dismiss the Past hypotheses and correctly  score the CF mode highest, underscoring its capability to discern semantic subtleties. 
Additionally, NLI's semantic generalization avoids the need for exact word matching, unlike the complex dictionary-based methods. For example, similar hypotheses such as ``increase'' and ``launch more protests'' receive similar scores, and phrases like ``promised to reduce'' are considered similar to ``will reduce''.

Our \textbf{mode-aware NLI} system leverages these insights, using the codebook to guide hypothesis construction. This approach enables easy conversion of present-tense label description into different modes or tenses, even for non-experts. 
The codebook also provides contrasting label examples, like ``YIELD: ease protests'' and  ``AGREE: agree to ease protests'' for the CP and CF modes for PROTEST, significantly  streamlining the engineering process.
We only need to ensure hypotheses, especially in Past mode, clearly reflect event trends.
For instance, in Table \ref{tab:nli mode examples},  ``protested against'' was rephrased  to ``increased/launched more protests against'' for enhanced clarity.  Likewise, ``imposed bans'' can be modified to ``increased/imposed more bans''.

\subsection{Class Disambiguation}\label{sec:class disambiguation}

To address the issue of class ambiguity and overlaps in CAMEO/PLOVER, experts have documented instructions and annotation rules in the codebook. Annotators frequently consult the codebook when faced with ambiguous cases. In contrast, we can integrate this information into our machine to reduce  manual annotation and effectively handling boundary cases.
Note that incorporating excessive rules goes against our goal of designing a simple and adaptable system.   It can lead to overfitting and inflexibility, similar to the limitations found in traditional dictionary-based methods. Therefore, we've chosen to include only the most frequent rules explicitly outlined in the codebook, considering this step as supplementary to our system.

One notable rule, referred to as the \textbf{Conflict Override}, is summarized from the codebook. This rule gives priority to labels in Material Conflict over Verbal Conflict, as depicted in Figure \ref{fig:framework}. 
If the top predictions include candidate labels in Material Conflict, the labels in Verbal Conflict will be overridden. 
For example,  we label ``protest to request'' as material PROTEST other than verbal REQUEST, as explained in Section \ref{sec:nli}.
Similarly, we label ``convict and arrest'' as material COERCE other than verbal ACCUSE, considering the more severe actions involved.
 These rules can be easily customized and expanded by users to accommodate changes in the schema or ontology. Additional discussions are provided in  Appendix \ref{appendix:mode table} and \ref{appendix:flexibility}.

\subsection{Tree-Query NLI Framework}
We enhance precision and efficiency by integrating mode-aware NLI with class disambiguation into a tree-query framework, as shown in Figure \ref{fig:framework}.   This contrasts with the ``flat-query'' method, where hypotheses are arranged at a single level, by organizing them hierarchically.

At \textbf{Level 1 Context}, we compare 76 Past hypotheses ($\approx 5$ hypotheses per Rootcode) to classify the premise's context.  Using a customized threshold, such as selecting the top-3 candidates with scores higher than the maximum score minus 0.1, we narrow down the most probable candidates. In the example, this filtering yields two candidates related to REQUEST and PROTEST.

At \textbf{Level 2 Mode}, we compare the hypotheses in other modes for the selected candidates to determine their mode. We focus on two types of modes in the experiments: Past and Future. For instance, PROTEST leads to two branches - the existing PROTEST and a new THREATEN (PROTEST+future). However, for certain Rootcodes like REQUEST, querying their Future variants is unnecessary since the labels remain the same from Past to Future (details in Table \ref{tab:plover_mode} and Appendix \ref{appendix:mode table}).  
This reduces the number of Future hypotheses in Level 2 to 58, and only a subset requires querying per premise.  
In Figure \ref{fig:framework}, we collect all necessary scores for Level 2 analysis by querying just the new THREATEN hypothesis once.

At \textbf{Level 3 Class Disambiguation}, we apply specific rules, including the Conflict Override, to eliminate REQUEST since PROTEST already exists among the top predictions from Level 2.

ZSP is interpretable, flexible, efficient, and precise. 
First,  we split the complicated, ambiguous classification into a simple tree framework that both computer science and political science researchers can easily understand. 
Second, experts can quickly update ZSP by revising the hypothesis table or class disambiguation rules according to a evolving ontology, which is much cheaper than maintaining large dictionaries or relabeling a dataset (detailed in Appendix \ref{appendix:flexibility}).
Third,  it improves efficiency. For instance, we only query 76 times in Level 1 + one time in Level 2 without comparing all 134 hypotheses in Figure \ref{fig:framework}. 
Finally, NLI scores within ZSP accurately capture nuanced entailment relations within the limited scope of compared hypotheses at each level. This minimizes potential errors that can arise from mixed hypotheses in different contexts and modes. We will validate this in experiments.

\subsection{ChatGPT}
Besides our proposed NLI-based ZSP model, we explored the zero-shot performance of LLMs on this task. We focused on two versions of ChatGPT: GPT-3.5 and GPT-4. We used the OpenAI API and designed prompts that incorporate task descriptions and pre-defined label sets, building upon insights from previous research \cite{Wei2023zero, Li2023evaluating}. The label descriptions were summarized and refined from the PLOVER codebook's comprehensive Rootcode descriptions. 
Further insights are available through exemplified input output instances in Appendix \ref{appendix:chatgpt}.

\section{Experiments}

\begin{figure*}
    \centering
    \subfloat[PLV Binary
    \label{subfig:plover_binary_different_size}]{
    \includegraphics[width=0.5\columnwidth,keepaspectratio]
    {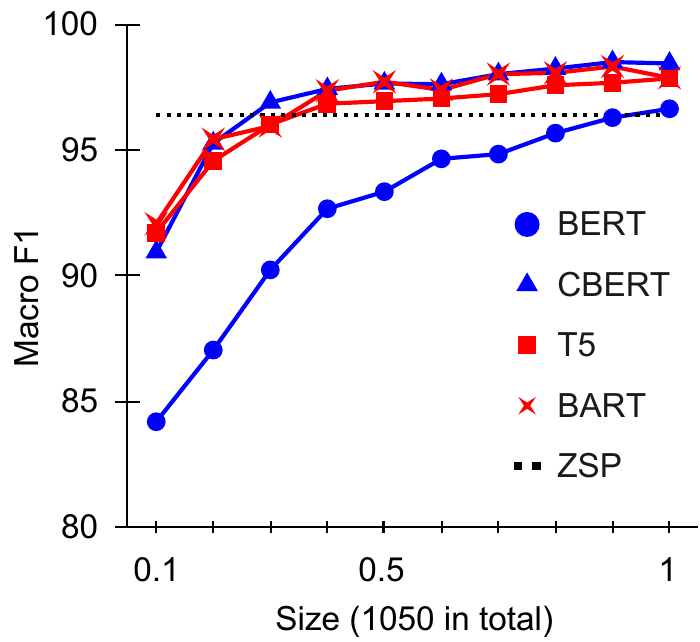}
    }
    \subfloat[PLV Quadcode
    \label{subfig:plover_quadcode_different_size}]{
    \includegraphics[width=0.5\columnwidth,keepaspectratio]{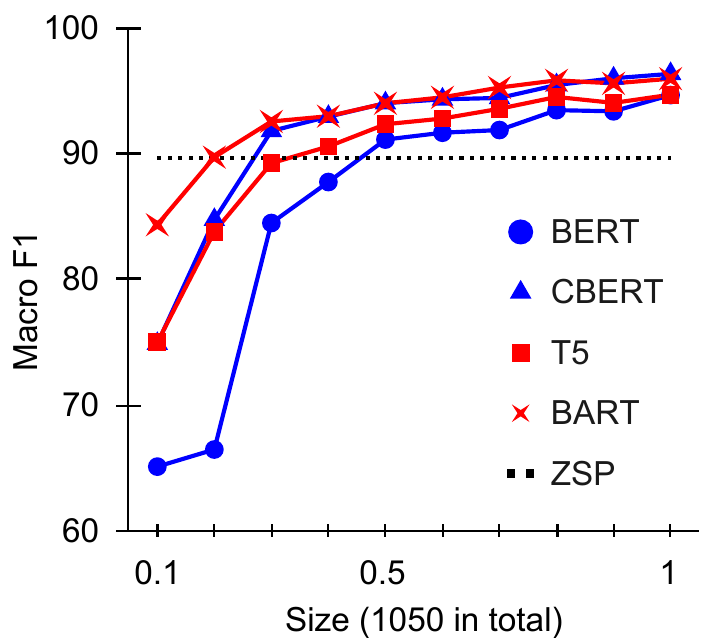}
    }
    \subfloat[PLV Rootcode
    \label{subfig:plover_rootcode_different_size}]{
    \includegraphics[width=0.5\columnwidth,keepaspectratio]{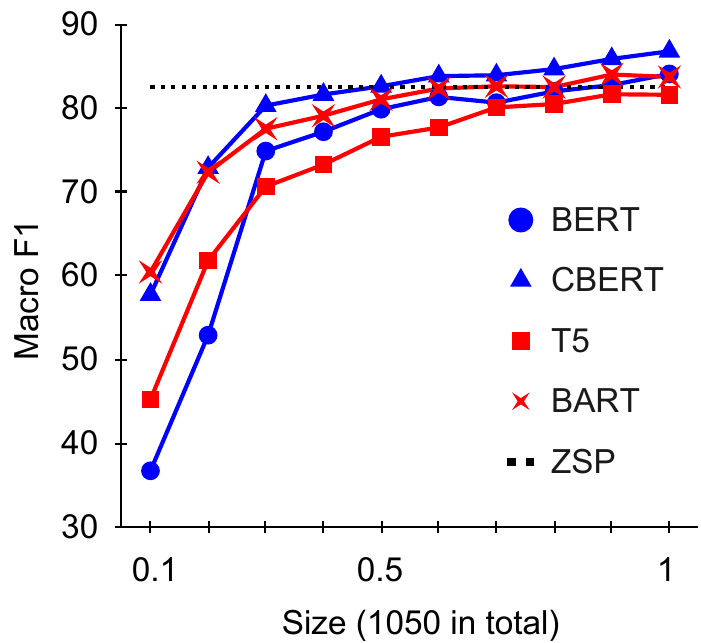}
    }
    \subfloat[A/W Binary
    \label{subfig:a/w_different_size}]{
    \includegraphics[width=0.5\columnwidth,keepaspectratio]{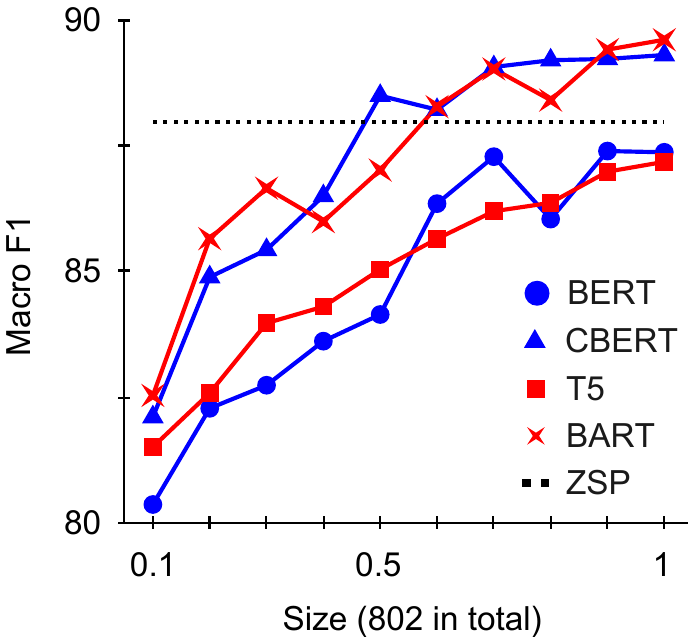}
    }
    \caption{Performance vs. varying sized training datasets.} \label{fig:different_size}

\end{figure*}

\subsection{Datasets}\label{sec:A/W datasets}

Since there were limited datasets with fine-grained annotation, we built a Rootcode-level \textbf{PLV} dataset from the CAMEO codebook and a balanced coarse-grained-labeled dataset \cite{parolin2022CoPED}, resulting in 1050 training examples and 1033 testing examples. We built three classification tasks with varying degrees of complexity: Binary (cooperation vs. conflict), Quadcode, and Rootcode. 

Besides the political science dataset PLV, we also explored how event ontology knowledge benefits and generalizes in other NLP datasets. 
Thus, we built a binary \textbf{A/W} dataset from \textbf{A}CE \cite{ace04} and \textbf{W}ikiEvents \cite{wikievent}, which contain many conflict-related subjects that overlap the political ontologies. A/W consists of 802 training examples and 805 testing examples.  See more details in Appendix \ref{appendix:aw}.

\subsection{Setup}
Regarding our proposed \textbf{ZSP}, we incorporated a finetuned NLI model\footnote{\url{https://huggingface.co/roberta-large-mnli}} into our tree-query system.
For \textbf{ChatGPT}, we used OpenAI's Chat completions API to access \textbf{GPT-3.5} and \textbf{GPT-4}.
To assess the practical usefulness of these zero-shot models, we compared them with notable baselines, including Universal PETRARCH (\textbf{UP}) \cite{UPetrarch}, a widely-used dictionary-based CAMEO event coder. We measured UP's ideal performance on relation classification by considering incomplete triplets, as detailed in Appendix \ref{appendix:up}.

Additionally, we examine the performance of various supervised learning models, including masking language models (MLM) like \textbf{BERT}-base-uncased \cite{bert} and ConfliBERT-scr-uncased (\textbf{CBERT}) \cite{conflibert2022}. Notably, CBERT reports greater effectiveness in the political science domain. 
We also use  text generation models, namely \textbf{BART} \cite{bart} and \textbf{T5} \cite{T5}, to generate original label texts for this classification task. 
We trained these supervised models on either the entire training set or sampled subsets of varying sizes using a single V-100 GPU with default hyperparameters. Subsequently, we  evaluated them on the complete testing dataset. 
We ran each scenario with five different seeds and reported average results for reliability.

\subsection{Results and Analysis}
\label{sec:Results}

We summarized the performance of dictionary-based and zero-shot models, as well as the supervised learning models trained on the entire training datasets, in Table \ref{tab:results summary}. Additionally, in Figure \ref{fig:different_size}, we compared ZSP with supervised learning models trained on varying limited datasets. UP and ChatGPT were excluded from the analysis due to their significant performance gap compared to the other models, to maintain focus and relevance.

\begin{table}\setlength{\tabcolsep}{4pt}\small
\centering
\begin{tabular}{ccccccc} 
\toprule
\textbf{Type}    & \textbf{Model} & \begin{tabular}[c]{@{}c@{}}\textbf{PLV }\\\textbf{Bin.}\end{tabular} & \begin{tabular}[c]{@{}c@{}}\textbf{PLV}\\\textbf{Quad}\end{tabular} & \begin{tabular}[c]{@{}c@{}}\textbf{PLV}\\\textbf{Root}\end{tabular} & \begin{tabular}[c]{@{}c@{}}\textbf{A/W}\\\textbf{Bin.}\end{tabular} & \textbf{Avg.}  \\ 
\midrule

\multirow{4}{*}{\begin{tabular}[c]{@{}c@{}}Dict. \&\\Zero-shot\end{tabular}} & UP    & 80.8    & 51.8    & 46.3    & 67.2    & 61.5    \\
    & GPT-3.5    & 90.1    & 66.2    & 40.9    & 76.3    & 68.4    \\
    & GPT-4  & 93.4 & 76.7 & 61.5 & 87.0 & 79.7  \\
    & ZSP    & \textbf{96.4}    & \textbf{89.6}    & \textbf{82.4}    & \textbf{88.0}    & \textbf{89.1}  \\ 
\midrule
\multirow{4}{*}{\begin{tabular}[c]{@{}c@{}}Super-\\vised\end{tabular}}    & BERT    & 96.6    & 94.6    & 84.0    & 87.4    & 90.7    \\
    & CBERT    & \textbf{98.4}    & \textbf{96.3}    & \textbf{86.7}    & 89.3    & \textbf{92.7}  \\
    & T5    & 97.8    & 94.7    & 81.6    & 87.2    & 90.3    \\
    & BART    & 97.9    & 95.9    & 83.7    & \textbf{89.6}    & 91.8    \\
\bottomrule
\end{tabular}
   \caption{Macro F1 scores of models on diverse dataset-task combinations and average results. }\label{tab:results summary}
\end{table}

\paragraph{Supervised learning.}

Among supervised models, CBERT emerged as the top performer, surpassing BERT with less data required.  BART closely trailed. It outperformed T5 by exhibiting less overfitting on small, imbalanced labeled datasets.

\paragraph{ZSP.} 
ZSP consistently outperformed UP and ChatGPT, and it achieved competitive results with supervised learning models in most tasks (Figures \ref{subfig:plover_binary_different_size}, \ref{subfig:plover_rootcode_different_size}, \ref{subfig:a/w_different_size}). Notably, in these scenarios, ZSP matched BERT and T5, while the stronger models CBERT and BART still required 25\%-50\% of the training data to achieve a slight performance gap (less than 4.3\%) over  ZSP.
The only exception was a notable 6.7\% performance gap observed between CBERT and ZSP on PLV-Quadcode (Figure \ref{subfig:plover_quadcode_different_size}). This difference can be attributed to the dataset's balanced and coarser-grained nature, which favors supervised learning.

However, supervised models experience a significant performance decline in more challenging fine-grained Rootcode classification (Figure \ref{subfig:plover_rootcode_different_size}), emphasizing the need for sufficient and balanced annotation.  
Actually, our experience across multiple projects to develop event coding datasets with approximately 1,000 examples typically extends beyond several months. Creating evaluation datasets like PLV and AW, or even relabeling existing ones, proves to be far more time-consuming.
In contrast, designing NLI prompts from the codebook for ZSP takes just a few days, greatly reducing annotation efforts and demonstrating clear advantages in real-world applications. Furthermore, ZSP’s lack of a training phase significantly cuts down on GPU resource needs, enhancing its adaptability and enabling efficient inference on both CPUs and GPUs. This efficiency starkly contrasts with supervised models, which rely heavily on costly GPU resources for training.

We further analyzed ZSP's confusion matrix for Rootcode classification (see Figure \ref{fig:confusion matrix on PLV Rootcode} in Appendix \ref{appendix:detailed result}). The results reveal high ZSP accuracy by correctly classifying most Rootcodes, yet there are some misclassifications, particularly for AGREE, SUPPORT, AID, and YIELD labels. 
These labels have subtle semantic differences, with AGREE representing a future, verbal, or hypothetical version of the other three categories.
For instance, consider the sentence labeled as diplomatic SUPPORT ``... \textless{}S\textgreater{} had approved an agreement with \textless{}T\textgreater{} ...'', ZSP produces conflicting predictions, with a score of 96.9\% for the hypothesis ``SUPPORT: approved an agreement'' and 97.0\% for ``AGREE: agreed to sign an agreement''. This discrepancy arises due to the fine distinction between these two labels, which even human annotators may find challenging.

\paragraph{ChatGPT.} 
We observed notable differences in the performance of GPT-3.5 and the latest GPT-4 models. Specifically, GPT-3.5 exhibited inconsistent results. Despite excelling in binary tasks, it struggles with more specific labels and even performs worse than UP in Rootcode classification. These challenges align with previous research in similar tasks \cite{yuan2023zero, Cai2023monte, Li2023evaluating, Gao2023exploring}. 

One ongoing challenge is generating formatted results and avoiding random labels outside the predefined set. To address this, we found that instructing GPT-3.5 to output digits (01-15) instead of text labels (AGREE - ASSAULT)  partially alleviates these challenges and improves recall scores.

Another difficulty lies in effectively incorporating complex task descriptions and predefined label information into GPT-3.5.  While our ZSP model can utilize class disambiguation rules easily, GPT-3.5 struggles to retain large amounts of information and may forget relevant details after just one round of chatting. 
This limitation necessitates the repetitive input of essential information in every interaction, which reduces efficiency. 

Furthermore, balancing the preservation of necessary information and the compression of prompts to accommodate actual questions proves challenging.
Continuous refinement of the prompts does not consistently improve performance, and it is counter-intuitive that longer label descriptions with more disambiguation instructions result in performance decline. The quest for an optimal prompt design remains an open question for future research.

However, GPT-4 stands out as a significant improvement over GPT-3.5.  It effectively reduces formatting errors, with few occasional issues lingering. The most significant enhancement is its ability to comprehend and process longer input tokens, allowing for better use of input information and finer class distinctions.  Interestingly, class disambiguation notes were found to be effective for GPT-4 but not for GPT-3.5, further distinguishing the two models. The success of GPT-4 highlights the vast potential of LLMs.  While extensive API queries can be costly, and precision may be slightly lower than ZSP, GPT-4's effectiveness with fewer prompts and superior generalization are notable advantages for future applications.

\begin{table}\setlength{\tabcolsep}{5pt}\small
\centering
\begin{tabular}{clccccc} 
\toprule
\multicolumn{2}{c}{\textbf{Model}}  & \begin{tabular}[c]{@{}c@{}}\textbf{PLV }\\\textbf{Bin.}\end{tabular} & \begin{tabular}[c]{@{}c@{}}\textbf{PLV}\\\textbf{Quad}\end{tabular} & \begin{tabular}[c]{@{}c@{}}\textbf{PLV}\\\textbf{Root}\end{tabular} & \begin{tabular}[c]{@{}c@{}}\textbf{A/W}\\\textbf{Bin.}\end{tabular} & \textbf{Avg.}  \\ 
\midrule
\multicolumn{2}{c}{UP}    & 80.8 & 51.8 & 46.3 & 67.2 & 61.5  \\
\multicolumn{2}{c}{GPT-3.5}  & 90.1 & 66.2 & 40.9 & 76.3 & 68.4  \\
\multicolumn{2}{c}{GPT-4}  & 93.4 & 76.7 & 61.5 & 87.0 & 79.7  \\

\midrule
\multirow{2}{*}{\begin{tabular}[c]{@{}c@{}}ZSP\\Flat \end{tabular}}  
    & Tiny    & 90.5    & 69.5    & 50.8    & 83.6 & 73.6 \\
    & Full    & 91.0    & 73.4    & 55.7    & 82.4 & 75.6 \\
    \midrule
\multirow{3}{*}{\begin{tabular}[c]{@{}c@{}}ZSP\\Tree\end{tabular}}    
    & $l_1$    & 96.2    & 85.8    & 78.2    & 87.8 & 87.0 \\
    & $l_{1,2}$     & \textbf{96.5}    & 87.6    & 79.4    & 87.8 & 87.8 \\
    & $l_{1,2,3}$     & 96.4    & \textbf{89.6}    & \textbf{82.4}    & \textbf{88.0} & \textbf{89.1} \\
\bottomrule
\end{tabular}
\caption{Macro F1 scores\% of ZSP with different settings vs. other zero-shot models in ablation study.}\label{tab:ablation study}
\end{table}

\subsection{Ablation Study}\label{sec:ablation study}

We conducted an ablation study to address the following questions on ZSP:
(1) Is a tree-query approach superior to a flat-query approach, which compares all hypotheses at single level simultaneously?
(2) Does having more hypotheses guarantee better performance?

Table \ref{tab:ablation study} displays the results of other zero-shot models, UP, GPT-3.5/4, and two variants of our ZSP models across multiple tasks.
For the \textbf{Flat}-query approaches, the \textbf{Tiny} model uses 18 hypotheses derived from the Rootcode names (See Table \ref{tab:preliminaries}).
The \textbf{Full} model incorporates a complete list of 222 label descriptions from the codebook.
The \textbf{Tree}-query approach consists of our ZSP model at different levels: $l_1$, $l_2$, and $l_3$.

The observation that the Tiny model with 18 hypotheses outperforms UP with $81k$ inflexible patterns, confirms the effectiveness of generalized PLM features.  
Furthermore, Tiny surpasses GPT-3.5, highlighting the unreliability of GPT-3.5 and emphasizing the significance of expert knowledge in achieving superior results.

Despite the Tiny model's limited capacity to handle nuanced cases, adding more unorganized hypotheses does not consistently improve performance. The Full model's improper mixing and comparison of hypotheses for verbal and material events at different levels result in arbitrary NLI scores, leading to poor performance on PLV and inferior results compared to the Tiny model on A/W.

In contrast, the tree-query models outperform all flat-query models by a large margin at Level 1. Adding additional levels brings stable improvements, primarily for Quadcode and Rootcode. The tree-query framework effectively delimits the scope of candidate hypotheses and offers precise NLI scores that capture semantic differences. This ensures a more controllable and accurate result.

\section{Conclusion} \label{sec:Conclusion}
Future event coding tools should prioritize ease of interpretation and flexibility, making them more practical than annotating new datasets for black-box supervised models. Therefore, we explored the potential of zero-shot relation classification using ChatGPT (GPT-3.5/4) and introduced our ZSP model.
While GPT-3.5 struggled with fine-grained classification, GPT-4 showed promise in mitigating instability issues. Our ZSP offers an even more cheap, precise, and adaptable solution. 
The key is structuring the complex problem into an interpretable, three-level tree framework, integrating mode-aware NLI, and incorporating class disambiguation rules from the codebooks. 
Overall, our study highlights the value of integrating transferred knowledge with expert linguistic insights to streamline the process of verifying event records for the political science community.

\section{Limitations} \label{sec:Limitations}

Balancing generalization and specificity is a common challenge across many methods. ZSP was developed to address the complexities of annotation codebooks and the inefficiencies in training annotators. This led to streamlined annotation, such as labeling an event as PROTEST instead of DEMAND for a protest related to rights, aided by the Conflict Override rule which simplifies complex annotation notes into machine-understandable rules. To maintain a balance between complexity and adaptability, we included only the most frequently used rules from the codebook.

Our approach's broader applicability, particularly in political science and related fields where codebooks are traditionally used to train annotators \cite{acled,gunviolence,parolin20213m}, underscores the practicality of our method in streamlining data labeling.   We are optimistic that combining established codebooks or knowledgebases with language models can extend to other domain-specific data. For instance, in legal studies, this approach can enhance the classification of legal documents by leveraging codified laws and regulations. In healthcare, it can aid in categorizing patient records and medical literature using clinical codebooks. In media studies and communication, it can support media content analysis by categorizing articles, broadcasts, and social media posts using thematic or sentiment-related codebooks.

However, challenges persist in zero-shot models when classifying semantically non-mutually exclusive fine-grained labels due to the intensive hypothesis engineering required. We addressed these challenges through the codebook's attainable expertise, but ZSP may struggle with tasks lacking accessible domain knowledge bases or those with overly nuanced and ambiguous labels. For example, classifying subcategories of ASSAULT (crime vs. attack vs. kidnap) or distinguishing peace protests from riots may require as many hypotheses as keywords \cite{barker-etal-2021-ibm,radford-2021-case}.
For such tasks, hybrid methods such as integrating ZSP or ChatGPT with few-shot learning, pattern-matching, or in-context learning could effectively address tasks of varying complexity, reducing human efforts. Future work will focus on exploring these hybrid methods.

Our exploration of ChatGPT models highlights the trade-off between generalizability and precision. While ChatGPT can adapt to any task with its chat-style format, it may sacrifice precision compared to NLI models. Due to time constraints and cost considerations, we did not investigate multi-turn interactions to enhance ChatGPT's precision, leaving this for future research. Nonetheless, both zero-shot models show promising potential to surpass traditional dictionary-based methods and annotation-driven supervised learning.

In selecting comparative methods, we included the most pertinent and recent ones that meet the specific needs of our task. Political event coding differs significantly from NLP event extraction tasks, such as those using the popular ACE dataset, in the availability of directly comparable studies. The prevailing methodologies in ACE event extraction are predominantly supervised and don't easily align with our unique ontology. Our investigation focuses on exploring the effectiveness of zero-shot models in event coding rather than achieving the highest accuracy through supervised approaches. Adapting existing zero-shot methods, designed with ACE contexts in mind, to our distinct requirements presents distinct challenges. Moving forward, we plan to expand our baseline comparisons in future studies.

While we initially considered expanding experiments to include other ontologies, we chose to focus on CAMEO/PLOVER, deferring broader explorations to future studies. This decision was influenced by practical constraints such as time and API costs, as well as a desire to pioneer within less-explored research domains. Unlike many widely-studied ontologies that rely on manually-created NLI systems and lack mode considerations, CAMEO/PLOVER presents unique challenges and opportunities. Its integration of mode features and codebooks makes it an ideal candidate for exploring PLMs in complex areas like political event coding. By converting the complex expertise embedded in the codebooks into practical applications, we transcend the limits of conventional zero-shot modeling and showcase how PLMs like NLI and ChatGPT can be adapted to specialized domains.

\section{Ethics Statement}
The broad goal of producing accurate event data is to objectively measure and understand processes of political conflict and mediation around the world in order to prevent or mitigate their harm. We aim to produce a simple, flexible tool to serve this purpose. In particular, the zero-shot approach of this study largely reduces the costs, effort, and time to produce highly-quality event data on conflict, thus helping international and domestic government agencies, as well as researchers and practitioners to track, analyze, and mitigate the causes and effects of political violence. 
The project relied exclusively on news-story-like text as second-hand accounts of conflict events, but did not involve human research subjects.

\section*{Acknowledgments}

The research reported herein was supported in part by NIST Award \# 60NANB23D007, NSF awards DMS-1737978, DGE-2039542, OAC-1828467, OAC-1931541, OAC-2311142, and DGE-1906630, ONR awards N00014-17-1-2995 and N00014-20-1-2738, and the National Center for Transportation Cybersecurity and Resiliency (TraCR).

\bibliography{reference}
\bibliographystyle{acl_natbib}

\clearpage

\appendix

\section{Mode Design and Mapping}\label{appendix:PLOVER2mode_mapping}

PLOVER suggests auxiliary modes to indicate whether a reported event is historical, future-oriented, hypothetical, or negated, as shown in Table \ref{tab:plover auxiliary modes}. Some event types can theoretically combine with an auxiliary mode, such as AGREE becoming SUPPORT + future or THREATEN becoming ASSAULT + hypothetical.  
However, PLOVER's guidance lacks concrete implementation for annotators,  merely assuming that ``the coding engine will be able to resolve these and put that information in the context.''

\begin{table}[b]\setlength{\tabcolsep}{2pt}\small
\centering
\resizebox{\columnwidth}{!}{
\begin{tabular}{lp{5.5cm}}
\toprule
\textbf{Mode} & \textbf{Example} \\
\midrule
Historical & During the decolonization struggle, Angolan forces...\\
Future & Members of the G-7 will meet in Ottawa next month...\\
Hypothetical & If Russian forces were to cross the border, that would represent a major...\\
Negation & Thus far, fighting has not re-emerged in the tense region.\\
\bottomrule
\end{tabular}}
\caption{Examples of PLOVER's auxiliary modes.}\label{tab:plover auxiliary modes}
\label{tab:event_modes}
\end{table}

\begin{table}\setlength{\tabcolsep}{2pt}\small
   \centering
   \scalebox{0.85}{
\begin{tabular}{rrrrrrrrrrr} 
\toprule
\textbf{\textbf{P}} &    &    & \textbf{\textbf{F}}  &    &    & \textbf{\textbf{CP}} &    &    & \textbf{\textbf{CF}} & \multicolumn{1}{l}{}  \\ 
\midrule
AGREE    & \multirow{3}{*}{1} & \multirow{3}{*}{} & \multirow{3}{*}{AGREE}    & \multirow{3}{*}{1} & \multirow{3}{*}{} & \multirow{3}{*}{REJECT}    & \multirow{3}{*}{3} & \multirow{3}{*}{} & \multirow{3}{*}{REJECT}    & \multirow{3}{*}{3}    \\
CONSULT    &    &    &    &    &    &    &    &    &    &    \\
SUPPORT    &    &    &    &    &    &    &    &    &    &    \\ 
\midrule
COOPERATE    & \multirow{3}{*}{2} & \multirow{3}{*}{} & \multirow{3}{*}{AGREE}    & \multirow{3}{*}{1} & \multirow{3}{*}{} & \multirow{3}{*}{SANCTION}   & \multirow{3}{*}{4} & \multirow{3}{*}{} & \multirow{3}{*}{REJECT}    & \multirow{3}{*}{3}    \\
AID    &    &    &    &    &    &    &    &    &    &    \\
YIELD    &    &    &    &    &    &    &    &    &    &    \\ 
\midrule
ACCUSE    & \multirow{4}{*}{3} & \multirow{4}{*}{} & ACCUSE    & \multirow{4}{*}{3} & \multirow{4}{*}{} & \multirow{4}{*}{AGREE}    & \multirow{4}{*}{1} & \multirow{4}{*}{} & \multirow{4}{*}{AGREE}    & \multirow{4}{*}{1}    \\
DEMAND    &    &    & DEMAND    &    &    &    &    &    &    &    \\
REJECT    &    &    & REJECT    &    &    &    &    &    &    &    \\
THREATEN    &    &    & THREATEN    &    &    &    &    &    &    &    \\ 
\midrule
PROTEST    & \multirow{5}{*}{4} & \multirow{5}{*}{} & \multirow{5}{*}{THREATEN} & \multirow{5}{*}{3} & \multirow{5}{*}{} & \multirow{5}{*}{YIELD}    & \multirow{5}{*}{2} & \multirow{5}{*}{} & \multirow{5}{*}{AGREE}    & \multirow{5}{*}{1}    \\
MOBILIZE    &    &    &    &    &    &    &    &    &    &    \\
SANCTION    &    &    &    &    &    &    &    &    &    &    \\
COERCE    &    &    &    &    &    &    &    &    &    &    \\
ASSAULT    &    &    &    &    &    &    &    &    &    &    \\
\bottomrule
\end{tabular}
   }
   \caption{PLOVER's labels (Rootcode text + Quadcode digits) w.r.t. our proposed modes: Past (\textbf{P}), Future (\textbf{F}),  Contradict\_Past (\textbf{CP}), Contradict\_Future (\textbf{CF}).}\label{tab:plover_mode}
\end{table}

Moreover,  while there are overlaps between PLOVER's auxiliary modes and the field of linguistic modality in NLP \cite{palmer2001mood,sauri2009factbank,rudinger2018neural,pyatkin2021possible}, notable differences exist. For instance, \citet{pyatkin2021possible} explore modes like event plausibility, which partially 
 echoes aspects of political actors' intentions  and event factuality in PLOVER. However, these explorations, though relevant, lack the precision and simplicity needed for direct application in PLOVER's context. 
Our focus, therefore, is on a simplified, practical, and task-specific mode framework for PLOVER.

Our proposed mode for PLOVER only consider four types: Past (\textbf{P}), Future (\textbf{F}),  Contradict\_Past (\textbf{CP}), Contradict\_Future (\textbf{CF}). These modes were derived from our examination of the CAMEO/PLOVER ontology and PLOVER's auxiliary modes from the PLOVER codebook. 
Within this framework, we make a clear distinction between verbal, future or hypothetical events (Future) and historical or ongoing events (Past).  And considering contradiction, we arrived at a simple 2x2 matrix with four modes outlined in Table \ref{tab:plover_mode}.
The table simplifies event coding and aids in accurately assigning Rootcode and Quadcode when an event's mode changes.

Specifically, Past covers historically significant or ongoing events, often presented in past tense but not restricted to it. 
Future includes verbal, hypothetical or future events. 
We consolidate hypothetical and future auxiliary modes in Table \ref{tab:event_modes} because their similar nature in transitions between material and verbal events.
For instance, THREATEN (Verbal Conflict, e.g., threatening to attack) can be considered either hypothetical or future ASSAULT (Material Conflict).
Contradict\_Past and Contradict\_Future encompass events contradicting Past or Future occurrences, respectively. As illustrated in Table \ref{tab:nli mode examples}, CF and CP may include words with contradictory meanings, not necessarily containing negation words like ``do not.'' Here, NLI’s ability to identify negation allows us to focus on positive hypotheses with contradictory meanings, aligning with PLOVER's guideline to exclude negated events from datasets. Moreover, the codebook already provides mirrored hypotheses, eliminating the need for manual construction. For example, ``YIELD: reduced protest against'' is the CP of ``PROTEST: protested against.''

An additional observation in Table \ref{tab:plover_mode} is that verbal actions remain classified as verbal regardless of mode. In contrast, material actions are categorized differently based on their contradictory forms.  
For instance, the contradiction or negation of AGREE (e.g., ``didn't agree to help'') is always REJECT, Verbal Conflict.
However, for material actions (e.g., ``provided aid to''), its CP form (e.g., ``stopped providing aid to'') is SANCTION, Material Conflict, but its CF form (e.g., ``would stop aid to'') is REJECT, Verbal Conflict. 

In sum, our task-specific mode concept aligns with PLOVER's auxiliary modes but enhances PLOVER's functionality, providing a practical, clear, and unambiguous approach to event coding.

\section{ZSP's Hypotheses and Class Disambiguation Rules}
\label{appendix:mode table}

Table \ref{tab:mode-aware hypothese table} shows the mode-aware hypotheses used in our experiments. 
We selected a subset of label descriptions in different Rootcode and Quadcode from the CAMEO codebook and converted these sentences to Past and Future modes. Some of them do not need Future variants as their labels from Past to Future remain the same, following Table \ref{tab:plover_mode}.

\begin{table}\small
\centering
\begin{tabular}{lr} 
\toprule
\multicolumn{1}{c}{\textbf{Hypothesis}} & \textbf{Label}  \\ 
\midrule
 \textless{}S\textgreater{} increased forces in \textless{}T\textgreater{}.    & MOBILIZE 4    \\
    \multicolumn{1}{c}{$\uparrow$ \texttt{override}} &   \\
 \textless{}S\textgreater{} increased peace forces in \textless{}T\textgreater{}.   & AID 2    \\

 \midrule

 \textless{}S\textgreater{} retreated forces from \textless{}T\textgreater{}.    & YIELD 2   \\
    \multicolumn{1}{c}{$\uparrow$ \texttt{override}}  &  \\
 \textless{}S\textgreater{} retreated peace forces from \textless{}T\textgreater{}. & SANCTION 4   \\
\bottomrule
\end{tabular}
\caption{Examples of class disambiguation: We override forces if top predictions contain peace forces.}
\label{tab:peace}
\end{table}

Crafting disambiguation rules is a smaller part of our work compared to developing broad event modes. Event modes have resolved many cases, with custom rules providing fine-tuning. Our primary goal is to maintain a simple and generalizable model. Therefore, we only add commonly encountered rules like \textbf{Conflict Override}, which is prevalent in the CAMEO codebook and affects the coding for all conflict events.

\textbf{Peace Override.}  
As the second frequent case, classes related to  ``forces'' vary according to actions and entities. For example, sending peacekeeping forces/workers/observers indicates cooperation, while sending forces to attack/occupy stands for conflict.
Thus, we add hypotheses with ``peace forces'' distinct from normal ``forces'', as shown in Table \ref{tab:peace}.
Predictions with ``peace forces'' have higher priority. I.e., we override ``forces'' if the top predictions contain ``peace forces'' because the latter one is more specified and infrequent. 
This simple rule ensures high recall for general forces and high precision for peace forces.

\begin{table}\setlength{\tabcolsep}{6pt}\small
\centering
\begin{tabular}{clccccc} 
\toprule
\multicolumn{2}{c}{\textbf{Model}}  & \begin{tabular}[c]{@{}c@{}}\textbf{PLV }\\\textbf{Bin.}\end{tabular} & \begin{tabular}[c]{@{}c@{}}\textbf{PLV}\\\textbf{Quad}\end{tabular} & \begin{tabular}[c]{@{}c@{}}\textbf{PLV}\\\textbf{Root}\end{tabular} & \begin{tabular}[c]{@{}c@{}}\textbf{A/W}\\\textbf{Bin.}\end{tabular} & \textbf{Avg.}  \\ 
\midrule
\multirow{4}{*}{\begin{tabular}[c]{@{}c@{}}ZSP\\Flat \end{tabular}}  & Tiny-$c$     & 89.7    & 68.9    & 49.5    & 81.0 & 72.3 \\
    & Tiny+$c$    & 90.5    & 69.5    & 50.8    & 83.6 & 73.6 \\
    & Full-$c$    & 89.4    & 70.8    & 53.1    & 75.9 & 72.3 \\
    & Full+$c$    & 91.0    & 73.4    & 55.7    & 82.4 & 75.6 \\
    \midrule
\multirow{5}{*}{\begin{tabular}[c]{@{}c@{}}ZSP\\Tree\end{tabular}}    & $l_1$-$c$   & 95.6    & 85.1    & 77.3    & 85.4 & 85.9 \\
    & $l_{1}$+$c$    & 96.2    & 85.8    & 78.2    & 87.8 & 87.0 \\
    & $l_{1,2}$-$c$    & 96.0    & 87.0    & 78.7    & 85.5 & 86.8 \\
    & $l_{1,2}$+$c$    & \textbf{96.5}    & 87.6    & 79.4    & 87.8 & 87.8 \\
    & $l_{1,2,3}$-$c$    &95.9	&89.0	&81.8	&85.5	&88.1 \\
    & $l_{1,2,3}$+$c$    & 96.4    & \textbf{89.6}    & \textbf{82.4}    & \textbf{88.0} & \textbf{89.1} \\
\bottomrule
\end{tabular}
\caption{Supplementary ablation study for Table \ref{tab:ablation study}. Macro F1 scores\% of ZSP with ($+c$) or without ($-c$) Consult Penalty in different configurations.}\label{tab:ablation study for consult}
\end{table}

\textbf{Consult Penalty.} 
Another common issue found in CAMEO/PLOVER is the overly general CONSULT class (e.g., consult/talk/meet/visit).
Many actions (e.g., sending forces, attacks, and investigations) entail that the source visited  the target. Likewise, an accusation or threat indicates that the source talked or met with the target.
One simple solution is to deduct the Consult Penalty, denoted as $c$ (e.g., 2\%), which penalizes the predicted entailment scores for the Rootcode ``CONSULT''.

We  analyze the impact of $c$ at every level in Table \ref{tab:ablation study for consult},  with ($+c$)  indicating results with the penalty and ($-c$)  showing results without it.  
The effect of $c$ is evident, with an average increase of 1.6\% in macro F1 for all the tasks.  
For deeper levels, $c$ ensures the accuracy of Level 1 predictions to avoid error propagation. 
These findings confirm the importance of preventing overly general and ambiguous hypotheses. 
Incorporating $c$ provides a simple solution to alleviate  manual efforts in curating alternative hypotheses.	

Besides the three main disambiguation rules, users can easily add or tailor less-important rules for their specific study purposes, as discussed in  Appendix \ref{appendix:flexibility}.

\section{Flexibility}\label{appendix:flexibility}

The ZSP framework is notably flexible, easily accommodating changes in ontology or schema. Experts can swiftly update the ZSP method by modifying the hypothesis table or adjusting the class disambiguation rules to align with an evolving ontology. 
For example, if political scientists reclassify "arresting someone" from ACCUSE to COERCE (Table \ref{tab:plover_mode}), they need only update the hypothesis label for ``<S> arrested person of <T>''. 
Similarly, introducing sub-categories within YIELD involves simple updates to the hypothesis labels.

\begin{table}
\small
\centering
\begin{tabular}{|p{7.2cm}|}
\hline
\textbf{Rootcode: \hspace{1em} 14-PROTEST} \\ 
\textbf{Code 144:} \hspace{0.6em} \emph{Obstruct passage, block, not specified below} \\ 
\textbf{Description:} Protest by blocking entry/exit into a building or area. \\ 
\textbf{Usage Notes:} Use sub-categories if demands are known. Use this code for protests disrupting routine proceedings by blocking roads, buildings, etc. Use code 191 if the blockade involves military forces. \\ \hline
\textbf{Rootcode: \hspace{1em} 19-COERCE} \\ 
\textbf{Code 191:} \hspace{0.5em} \emph{Impose blockade, restrict movement} \\ 
\textbf{Description:} Prevent entry/exit from a territory using armed forces. \\ 
\textbf{Usage Notes:} Different from code 144, which refers to civilian protests. \\ \hline
\end{tabular}
\caption{Examples of CAMEO classes with nuanced differences. Customized rules can differentiate these classes based on the actors involved.}\label{tab:blockade override}
\end{table}

Disambiguation rules, such as the Conflict Override rule, which prioritizes PROTEST over REQUEST in certain contexts, can also be refined. Transitioning to a multi-label approach is straightforward by eliminating the Conflict Override rule to acknowledge both PROTEST and REQUEST as valid labels.

When should users write their own disambiguation rules?  The need for custom rules depends on specific user requirements and the balance between manual effort and system precision and recall.  Custom rules can be particularly beneficial for fine-grained analysis.  
For example, the CAMEO codebook includes similar classes ``COERCE- Impose blockade'' and ``PROTEST- Obstruct passage/ blockade'', as shown in Table \ref{tab:blockade override}. 
The key difference is whether the source is armed forces or protestors.

For researchers focusing on in-depth civil protest studies\footnote{\url{https://github.com/emerging-welfare/glocongold}.}, distinguishing between codes 144 (civilian protests) and 191 (military blockades) is crucial for accurate classification.  Thus, they can define a simple rule, \textbf{Blockade Override}, without additional cost: remove the hypothesis ``COERCE- Impose blockade'' if the top predictions contain PROTEST, indicating that the source is more likely protestors rather than armed forces. This adaptability showcases the model's flexibility and customizability in complex political scenarios.

While ChatGPT can be generalized to any task with their chat-style format, they may sacrifice precision compared with the ZSP model. Yet, both zero-shot models show promising applicability to surpass traditional dictionary-based methods or annotation-driven supervised learning methods.

\begin{table}\setlength{\tabcolsep}{3pt}\small
    \centering
    \begin{tabular}{crrrc} 
    \toprule
    \multicolumn{1}{l}{\textbf{Dataset}} & \textbf{Subset}    & \multicolumn{1}{c}{\textbf{\# Docs}} & \multicolumn{1}{l}{\textbf{\# S-T pairs}} & \textbf{Tasks}    \\ 
    \midrule
    \multirow{3}{*}{PLV}    & CoPED    & -    & 1043/698    & \multirow{3}{*}{\begin{tabular}[c]{@{}c@{}}Binary,\\Quadcode, \\Rootcode\end{tabular}}  \\
    & Codebook   & -    & 0/335    &    \\
    & Total    & -    & 1050/1033    &    \\ 
    \midrule
    \multirow{3}{*}{A/W}    & ACE    & 337/338    & 432/451    & \multirow{3}{*}{Binary}    \\
    & WikiEvents & 91/92    & 370/434    &    \\
    & Total    & 428/430    & 802/805    &    \\
    \bottomrule
    \end{tabular}
    \caption{Statistics of the datasets: subsets, No. of documents and source-target pairs, and train/test splits.}\label{tab:dataset statistics}
\end{table}

\begin{figure}\fontsize{8pt}{8pt}\
\centering
\begin{BVerbatim}[commandchars=\\\{\}]
\textbf{Conflict.attack}: <arg1:attacker> attacked <arg2:tar-
get> using <arg3:instrument> at <arg4:place> place.
\textbf{Justice.arrest}: <arg1:jailer> arrested <arg2:detain-
ee> for <arg3:crime> crime at <arg4:place> place.
\end{BVerbatim}  
    \caption{Examples of templates in the A/W's original ontology \cite{wikievent}.}
    \label{fig:AW event ontology example}
\end{figure}

\section{Building PLV and A/W Datasets}\label{appendix:aw}

\begin{figure}
    \centering
    \subfloat[Quadcode
    \label{subfig:PLV Quadcode statistics}]{
    \includegraphics[width=0.35\columnwidth,keepaspectratio]
    {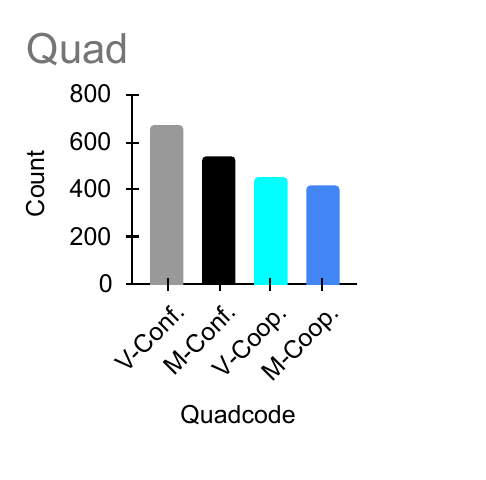}
    }
    \quad
    \subfloat[Rootcode
    \label{subfig:PLV Rootcode statistics}]{
    \includegraphics[width=0.45\columnwidth,keepaspectratio]{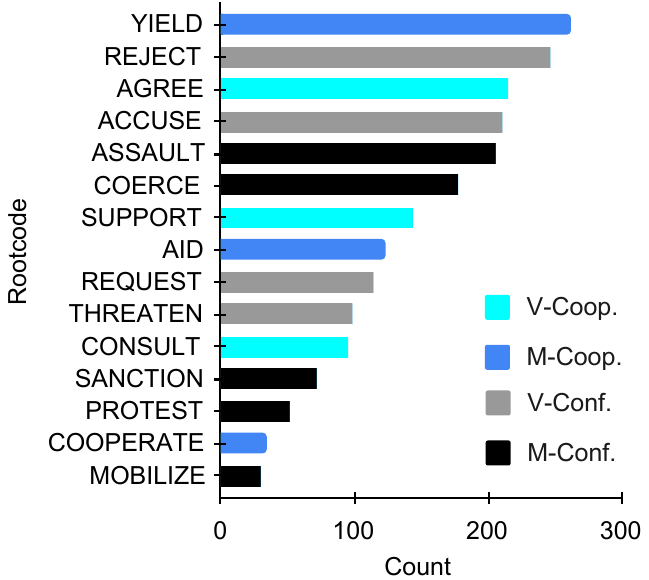}
    }
    \caption{Extending PLV-Quadcode to Rootcode level.} \label{fig:PLV dataset summary}
\end{figure}

We extended existing resources to build our datasets, which is more efficient and effective than creating a new dataset from scratch. 
Table \ref{tab:dataset statistics} summarizes the two datasets' detailed train and test split statistics. PLV is constructed from two resources. First, we outlined 335 examples (unique source-target pairs) with PLOVER Rootcode from the CAMEO \textbf{codebook}, and the PLOVER repository. Then we preprocessed a coarse-grained-labeled dataset from \textbf{CoPED} \cite{parolin2022CoPED} and manually extended its Quadcode labels to 15 Rootcode in the new PLOVER schema. The major modification can be seen in Table \ref{tab:preliminaries}. However, given that the current PLOVER codebook is in development, we leave YIELD without splitting it to CONCEDE and RETREAT. Finally, Figure \ref{fig:PLV dataset summary} visualizes our final dataset's label distribution.

We built the A/W dataset from the ACE and WikiEvents datasets. First, the repository of \cite{wikievent} provides templates for each event subtype of their ontology, enabling us to convert between different ontologies.
For example, Figure \ref{fig:AW event ontology example} shows two frequent event types defined in the ontology.
In both instances, argument 1 is equivalent to the source/actor, while argument 2 represents the target/recipient entities. Besides, the event type attack and arrest can be approximately mapped to ASSAULT and COERCE in PLOVER, respectively, as shown in Table \ref{tab:mapping A/W event types to plover category}. 

\begin{figure}
    \centering
    \subfloat[Event types statistics
    \label{subfig:aw event types}]{
    \includegraphics[width=0.65\columnwidth,keepaspectratio]{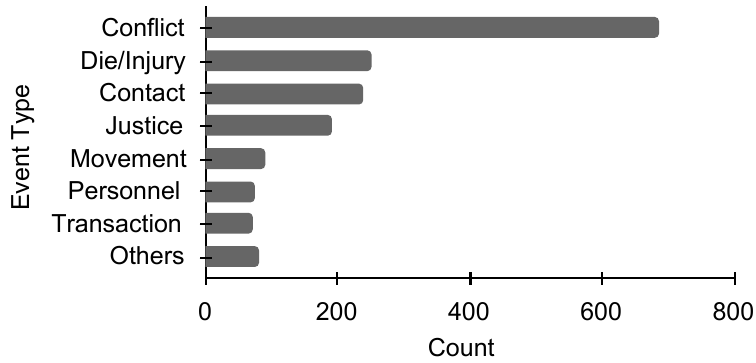}
    }
    \subfloat[Binary class
    \label{subfig:aw binary description}]{
    \includegraphics[width=0.25\columnwidth,keepaspectratio]{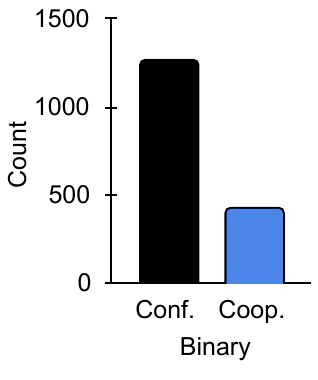}
    }
    \caption{A/W dataset's original event types and its relabeled binary category.} 
    \label{fig:aw dataset summary}
\end{figure}

Therefore, we built labeled source-target pairs from ACE and WikiEvents. We extracted major sentences that contained the labeled entities from each long document in WikiEvents. We also removed entities that only consist of pronouns. Finally, we got 1258 valid sentences with 1687 labeled Source-Target pairs.
To prevent label leaking, we split the dataset by document IDs, ensuring distinct name entities for training and testing. 
Figure \ref{fig:aw dataset summary} shows the distribution of the original event types and the mapped binary class.

The nuanced differences between the two domains necessitate that event types be only ``approximately'' mapped to PLOVER Rootcodes. And extensive manual verification is needed to ensure accuracy. This complexity is rooted in the distinct focuses of NLP, which emphasizes predicates or topic-centric events, and Political Science, which concentrates on event status or mode. For instance, examples in Tables \ref{tab:nli examples} (planned protest) and Table \ref{tab:nli mode examples} (agreement to suspend protests) are both categorized as Conflict.Demonstrate in A/W, but in PLOVER, they are distinctly classified as THREATEN (verbal conflict) and AGREE (verbal cooperation), respectively. The binary labels even switch from conflict to cooperation in the second case. Thus, manual checking remains crucial even at the binary level.

The annotation process was carried out by two authors, achieving a Kappa score of 0.76, with discrepancies resolved through discussion.

\section{UP Experiment Setup}\label{appendix:up}

Universal Petrarch (UP)  is a popular dictionary-based event coder \cite{UPetrarch}. We adapted UP into our task of relation classification with gold source and target, i.e., source-target-action triplets.
We found that UP is too strict and often results in incomplete or empty triplets. Thus, we reported the best possible result by the following methods.
First, we used UP for each sentence to extract all possible events.
Then we ranked the extracted triplets by the number of matched entities with gold sources and targets to decide the event code. 
We also counted the valid event code when there were no matched entities but only matched trigger action verbs.
Even so, there are still 10\% and 27\%  invalid event code results on PLV and A/W datasets, respectively.
Finally, we mapped its output four-digit code to PLOVER Rootcode and Quadcode (similar to Figure \ref{fig:cameo example}).

\begin{table}\setlength{\tabcolsep}{3pt}\small
    \centering
    \begin{tabular}{rrc} 
    \toprule
    \textbf{A/W Event Types}    & \textbf{Approx. Root.}   & \textbf{Binary}    \\ 
    \midrule
    Life.Die/Injure    & ASSAULT    & \multirow{4}{*}{Confli.}  \\
    Conflict.Attack    & ASSAULT    &    \\
    Conflict.Demonstrate    & PROTEST    &    \\
    Justice    & ACCUSE or COERCE &    \\
    \midrule
    Personnel.EndPosition & YIELD    & \multirow{4}{*}{Coop.~}   \\
    Contact    & CONSULT    &    \\
    Transaction    & COOPERATE or AID &    \\
    Business.Merge-Org    & COOPERATE~    &    \\
    \bottomrule
    \end{tabular}
    \caption{Mapping A/W's event types to PLOVER's approximate  Rootcode and binary class. }\label{tab:mapping A/W event types to plover category}
\end{table}

\section{ZSP's Detailed Results Analysis}\label{appendix:detailed result}

\begin{figure}[t]
    \centering
    \includegraphics[width=\linewidth]
    {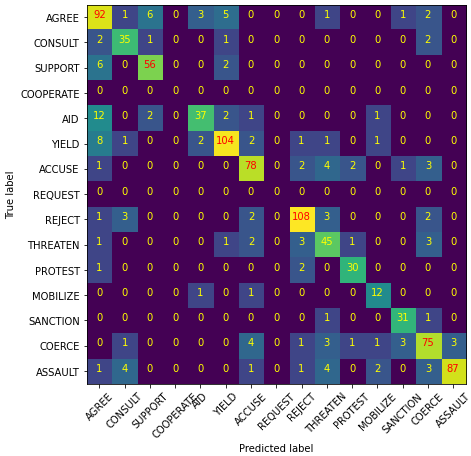}
    \caption{Confusion matrix for ZSP on PLV Rootcode.}
    \label{fig:confusion matrix on PLV Rootcode}
\end{figure}

\begin{figure}
    \begin{minipage}{0.45\linewidth}\centering
    \includegraphics[width=0.8\columnwidth,keepaspectratio,trim={0mm 0mm 0mm 7mm},clip]
    {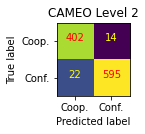}
    \caption{Confusion matrix for ZSP on PLV Binary code.}
    \label{fig:confusion matrix on plv binary}
    \end{minipage}%
    \hfill
    \begin{minipage}{0.5\linewidth}
    \includegraphics[width=\columnwidth,keepaspectratio,trim={0mm 0mm 0mm 8mm},clip]
    {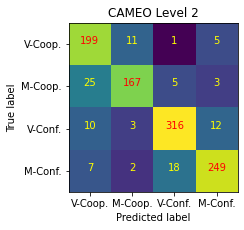}
    \caption{Confusion matrix for ZSP on PLV Quadcode.}
    \label{fig:confusion matrix on plv quadcode}
    \end{minipage} 
\end{figure}

We examined  the confusion matrix
for ZSP on Binary (Figure \ref{fig:confusion matrix on plv binary}), Quadcode (Figure \ref{fig:confusion matrix on plv quadcode}), and Rootcode (Figure \ref{fig:confusion matrix on PLV Rootcode}) classifications.
The results show that ZSP perfectly classifies most  contexts, with a slight degradation in differentiating 
 mode (verbal vs. material).

 In-depth class reports for PLV on Quadcode (Table \ref{tab:plv quadcode class report}) and Rootcode (Table \ref{tab:plv rootcode class report}) reveal that ZSP outperforms UP in nearly all metrics, except in the precision of the Verb-Conflict class (85.5\%). However, UP's lower recall impacts its overall F1 score, showcasing the superiority of PLM's generalized knowledge over rigid pattern-matching approaches. Additionally, we noticed  a performance trade-off when using overrides from Level 2 to Level 3. For instance, recall improves in Material-Conflict but decreases in Verbal-Conflict. Nevertheless, Level 3 significantly enhances overall F1 scores.

\begin{figure}
    \centering
    \includegraphics[width=\linewidth]
    {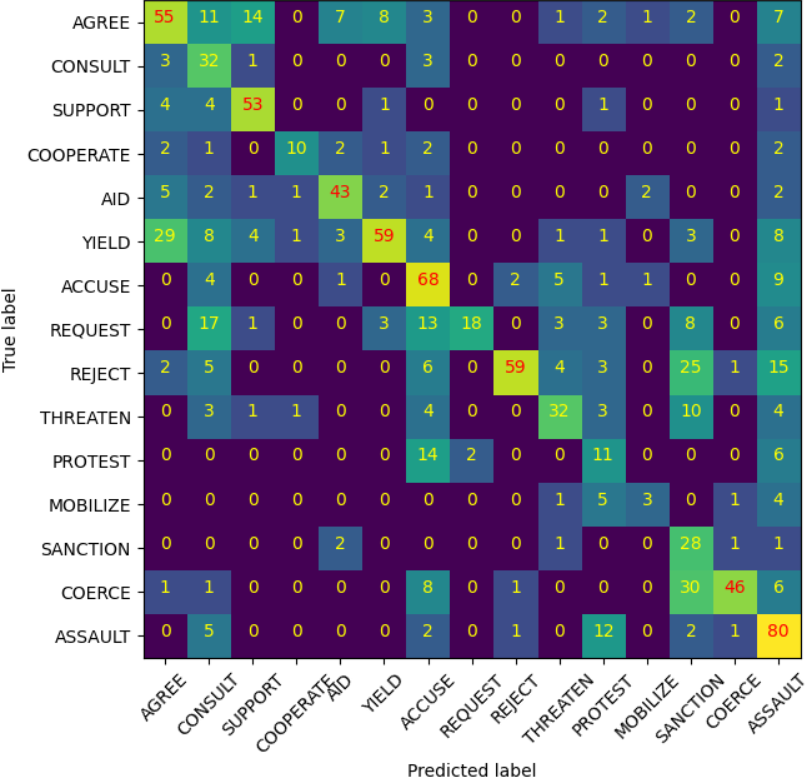}
    \caption{Confusion matrix on PLV Rootcode using the ``Full'' model in Section \ref{sec:ablation study} Ablation Study.}
    \label{fig:confusion matrix for ZSP Full model}
\end{figure}

Further, we expand on the ablation study (Section \ref{sec:ablation study}), emphasizing why the tree-query approach, with fewer hypotheses, surpasses the ``Full'' model, which utilizes 222 flat hypotheses.  
Figure \ref{fig:confusion matrix for ZSP Full model} illustrates the confusion matrix for the Full model's Rootcode classification. A comparison between this matrix (Figure \ref{fig:confusion matrix for ZSP Full model}) and the default ZSP model using tree-query (Figure \ref{fig:confusion matrix on PLV Rootcode}) reveals significant differences. The variable nature of NLI scores is a key factor in these differences. The tree-query model's focused approach on controlled hypothesis groups with consistent entities and predicates, but varying modes, leads to more accurate hypothesis identification. In contrast, the Full model's flat amalgamation of diverse hypotheses results in unpredictable outcomes and struggles with accurate mode classification, evident in frequent misclassifications between categories such as AGREE vs. SUPPORT, YIELD vs. AGREE, and REJECT vs. SANCTION or ASSAULT.

\section{NLI Model Selection}\label{appendix:roberta-base}

We selected RoBERTa-Large-MNLI\footnote{https://huggingface.co/roberta-large-mnli} for its extensive usage in NLI research, with comparable alternatives like BART-Large-MNLI\footnote{https://huggingface.co/facebook/bart-large-mnli} also showing favorable results. 
Employing smaller-sized base models for zero-shot tasks is less common, primarily due to the significant drop in performance.  Consequently, there are limited models specifically designed and widely accepted for zero-shot classification tasks.

\begin{table}\setlength{\tabcolsep}{5pt}\small
    \centering
    \begin{tabular}{ccccccc}
        \toprule
 
        \textbf{Model}  & \textbf{Size} & \begin{tabular}[c]{@{}c@{}}\textbf{PLV }\\\textbf{Bin.}\end{tabular} & \begin{tabular}[c]{@{}c@{}}\textbf{PLV}\\\textbf{Quad}\end{tabular} & \begin{tabular}[c]{@{}c@{}}\textbf{PLV}\\\textbf{Root}\end{tabular} & \begin{tabular}[c]{@{}c@{}}\textbf{A/W}\\\textbf{Bin.}\end{tabular} & \textbf{Avg.}  \\ 
        
        \midrule
        base & 125M & 95.2 & 83.0 & 68.4 & 81.1 & 81.9 \\
        large & 355M  & 96.4 & 89.6 & 82.4 & 88.0 & 89.1 \\
        \bottomrule
    \end{tabular}
    \caption{Macro F1 scores of ZSP models with different sized RoBERTa NLI models.}\label{tab:base model vs large model}
\end{table}

From an efficiency perspective, employing large models for zero-shot tasks proves efficient as they are only required during the inference phase. Conversely, training supervise large models can be relatively expensive. Besides,  one of our chosen baselines, CBERT \cite{conflibert2022}, only has a base version. Therefore, we conducted supervised experiments using base models while reserving large models exclusively for zero-shot tasks. This approach ensures a relatively fair and meaningful comparison between the two model types.

However, we also considered the possibility that a more rigorous comparison could have strengthened our hypotheses, particularly in demonstrating the effectiveness of smaller base models for handling fine-grained tasks in zero-shot scenarios.  
To explore this, we conducted experiments using an existing RoBERTa base model\footnote{https://huggingface.co/cross-encoder/nli-roberta-base}. The results are presented in Table \ref{tab:base model vs large model}, offering valuable additional insights alongside the findings presented in Table \ref{tab:results summary}.  While we observed that base models can effectively classify context or topics, they encountered challenges in distinguishing nuanced differences in mode. This distinction can lead to a drop in performance compared to larger models.

\begin{table}\setlength{\tabcolsep}{4pt}\small
    \centering
    \begin{tabular}{crccccc} 
    \toprule

\multirow{2}{*}{\textbf{Class}}    & \multirow{2}{*}{\textbf{No.}}  & \multirow{2}{*}{\textbf{Metrics}} & \multirow{2}{*}{\textbf{UP}} & \multicolumn{3}{c}{\textbf{ZSP}}    \\
    &    &    &    & $l_{1}$   &  $l_{1,2}$   & $l_{1,2,3}$    \\
    
    \midrule
    \multirow{3}{*}{V-Coop.}    & \multirow{3}{*}{216}  & Precison & 63.1  & \textbf{82.9} & \textbf{82.9} & 82.6    \\
    &    & Recall   & 68.1  & 83.3    & \textbf{92.1} & \textbf{92.1}  \\
    &    & Macro F1 & 65.5  & 83.1    & \textbf{87.3} & 87.1    \\ 
    \midrule
    \multirow{3}{*}{M-Coop.}    & \multirow{3}{*}{200}  & Precison & 52.4  & 84.1    & \textbf{91.7} & 91.3    \\
    &    & Recall   & 60.5  & \textbf{84.5} & 83.0    & 83.5    \\
    &    & Macro F1 & 56.1  & 84.3    & 87.1    & \textbf{87.2}  \\ 
    \midrule
    \multirow{3}{*}{V-Conf.}    & \multirow{3}{*}{341}  & Precison & 85.5  & 85.9    & 85.9    & \textbf{92.9}  \\
    &    & Recall   & 51.9  & \textbf{94.4} & \textbf{94.4} & 92.7    \\
    &    & Macro F1 & 64.6  & 89.9    & 89.9    & \textbf{92.8}  \\ 
    \midrule
    \multirow{3}{*}{M-Conf.}    & \multirow{3}{*}{276}  & Precison & 75.7  & 92.1    & \textbf{93.2} & 92.6    \\
    &    & Recall   & 69.9  & 80.1    & 80.1    & \textbf{90.2}  \\
    &    & Macro F1 & 72.7  & 85.7    & 86.2    & \textbf{91.4}  \\ 
    \midrule
    \multirow{3}{*}{\begin{tabular}[c]{@{}c@{}}macro \\avg.\end{tabular}} & \multirow{3}{*}{1033} & Precison & 55.3  & 86.2    & 88.4    & \textbf{89.8}  \\
    &    & Recall   & 50.1  & 85.6    & 87.4    & \textbf{89.6}  \\
    &    & Macro F1 & 51.8  & 85.8    & 87.6    & \textbf{89.6}  \\
    \bottomrule
    \end{tabular}
    \caption{PLV Quadcode performance analysis.}\label{tab:plv quadcode class report}
\end{table}

\section{ChatGPT Experiment Setup}\label{appendix:chatgpt}
Table \ref{tab:chatgpt} exemplifies inputs for relation classification tasks. 
Our task is characterized by challenging fine-grained classification that demands a substantial amount of input information. 
Due to token limitations and API costs, inputting one example at a time with a lengthy prompt is inefficient and costly. Instead, we used a long prompt followed by a list of input sentences to stay within the maximum token limits and obtain a list of predicted labels.
More specifically, 
the inputs comprise the task and label description, a sentence list (usually limited to less than 50 sentences due to word constraints), and the task requirements. The anticipated output from the model is the predicted labels. 
Despite our repeated emphasis on ChatGPT generating only predefined labels, certain issues remain. To mitigate these, we use numerical codes (01-15) instead of text labels (AGREE - ASSAULT), reducing ChatGPT's generation of labels outside the predefined set. Additionally, we've noticed that ChatGPT tends to forget the task description and predefined label information, necessitating their input each time. Finally, refining the task and label description doesn't yield improved results. This underscores the complexity of the task, involving semantically non-mutually exclusive fine-grained labels, which proves challenging for ChatGPT.

\begin{table}[t]\setlength{\tabcolsep}{4pt}\small
    \centering
    \begin{tabular}{ccccr}
    \toprule
    \textbf{Class}    & \textbf{Precision}   & \textbf{Recall}   & \textbf{Macro F1}    & \textbf{No.}  \\ 
    \midrule
    AGREE    & 73.6 & 82.9 & 78.0  & 111   \\
    CONSULT    & 70.0 & 85.4 & 76.9 & 41    \\
    SUPPORT    & 84.8 & 87.5 & 86.2 & 64    \\
    COOPERATE    & 68.2 & 75.0  & 71.4 & 20    \\
    AID    & 82.2 & 62.7 & 71.2 & 59    \\
    YIELD    & 89.7 & 86.0  & 87.8 & 121   \\
    ACCUSE    & 82.1 & 85.7 & 83.9 & 91    \\
    REQUEST    & 96.8 & 84.7 & 90.4 & 72    \\
    REJECT    & 90.8 & 90.0   & 90.4 & 120   \\
    THREATEN    & 71.4 & 77.6 & 74.4 & 58    \\
    PROTEST    & 88.2 & 90.9 & 89.6 & 33    \\
    MOBILIZE    & 66.7 & 85.7 & 75.0  & 14    \\
    SANCTION    & 86.1 & 93.9 & 89.9 & 33    \\
    COERCE    & 82.4 & 80.6 & 81.5 & 93    \\
    ASSAULT    & 96.7 & 84.5 & 90.2 & 103   \\
    \midrule
    accuracy    &    &    & 83.8 & 1033  \\
    macro-avg.   & 82.0  & 83.5 & 82.4 & 1033  \\
    \bottomrule
    \end{tabular}

    \caption{PLV Rootcode performance analysis.}
    \label{tab:plv rootcode class report}
\end{table}

\clearpage
\begin{table*}[htbp]
\centering
\resizebox{!}{0.46\textheight}{
\begin{tabular}{llll}
\toprule
\textbf{Root.}    & \textbf{Quad.} & \textbf{Past}    & \textbf{Future}    \\
\midrule
AGREE    & V-Coop.   & \textless{}S\textgreater{} agreed to do something for \textless{}T\textgreater{}    & None    \\
AGREE    & V-Coop.   & \textless{}S\textgreater{} promised to do something for \textless{}T\textgreater{}    & None    \\
CONSULT    & V-Coop.   & \textless{}S\textgreater{} held a talk with \textless{}T\textgreater{}    & \textless{}S\textgreater{} agreed to hold a talk with \textless{}T\textgreater{}    \\
CONSULT    & V-Coop.   & \textless{}S\textgreater{} met with \textless{}T\textgreater{}    & \textless{}S\textgreater{} agreed to meet with \textless{}T\textgreater{}    \\
CONSULT    & V-Coop.   & \textless{}S\textgreater{} undertook more negotiation with \textless{}T\textgreater{}    & \textless{}S\textgreater{} agreed to undertake negotiation with \textless{}T\textgreater{}    \\
SUPPORT    & V-Coop.   & \textless{}S\textgreater{} apologized to \textless{}T\textgreater{}    & \textless{}S\textgreater{} agreed to apologize to \textless{}T\textgreater{}    \\
SUPPORT    & V-Coop.   & \textless{}S\textgreater{} expressed support for \textless{}T\textgreater{}    & \textless{}S\textgreater{} agreed to support \textless{}T\textgreater{}    \\
SUPPORT    & V-Coop.   & \textless{}S\textgreater{} granted diplomatic recognition of \textless{}T\textgreater{}    & \textless{}S\textgreater{} agreed to grant diplomatic recognition of \textless{}T\textgreater{}    \\
SUPPORT    & V-Coop.   & \textless{}S\textgreater{} improved diplomatic cooperation with \textless{}T\textgreater{} & \textless{}S\textgreater{} agreed to improve diplomatic cooperation with \textless{}T\textgreater{}    \\
SUPPORT    & V-Coop.   & \textless{}S\textgreater{} signed an agreement with \textless{}T\textgreater{}    & \textless{}S\textgreater{} agreed to sign an agreement with \textless{}T\textgreater{}    \\
AID    & M-Coop.    & \textless{}S\textgreater{} added aid to \textless{}T\textgreater{}    & \textless{}S\textgreater{} agreed to provide aid to \textless{}T\textgreater{}    \\
AID    & M-Coop.    & \textless{}S\textgreater{} added money to \textless{}T\textgreater{}    & \textless{}S\textgreater{} agreed to add money to \textless{}T\textgreater{}    \\
AID    & M-Coop.    & \textless{}S\textgreater{} granted asylum to \textless{}T\textgreater{}    & \textless{}S\textgreater{} agreed to grant asylum to \textless{}T\textgreater{}    \\
AID    & M-Coop.    & \textless{}S\textgreater{} increased peace forces in \textless{}T\textgreater{}    & \textless{}S\textgreater{} agreed to increase peace forces in \textless{}T\textgreater{}    \\
COOPERATE & M-Coop.    & \textless{}S\textgreater{} cooperated with \textless{}T\textgreater{}    & \textless{}S\textgreater{} agreed to cooperate with \textless{}T\textgreater{}    \\
COOPERATE & M-Coop.    & \textless{}S\textgreater{} extradited person to \textless{}T\textgreater{}    & \textless{}S\textgreater{} agreed to extradite person to \textless{}T\textgreater{}    \\
COOPERATE & M-Coop.    & \textless{}S\textgreater{} shared information with \textless{}T\textgreater{}    & \textless{}S\textgreater{} agreed to share information with \textless{}T\textgreater{}    \\
YIELD    & M-Coop.    & \textless{}S\textgreater{} accepted demands of \textless{}T\textgreater{}    & \textless{}S\textgreater{} promised to accept demands of \textless{}T\textgreater{}    \\
YIELD    & M-Coop.    & \textless{}S\textgreater{} allowed entry of \textless{}T\textgreater{}    & \textless{}S\textgreater{} promised to allow entry of \textless{}T\textgreater{}    \\
YIELD    & M-Coop.    & \textless{}S\textgreater{} declared a ceasefire with \textless{}T\textgreater{}    & \textless{}S\textgreater{} promised to a ceasefire with \textless{}T\textgreater{}    \\
YIELD    & M-Coop.    & \textless{}S\textgreater{} eased restrictions on \textless{}T\textgreater{}    & \textless{}S\textgreater{} promised to ease restrictions on \textless{}T\textgreater{}    \\
YIELD    & M-Coop.    & \textless{}S\textgreater{} provided rights to \textless{}T\textgreater{}    & \textless{}S\textgreater{} promised to provide rights to \textless{}T\textgreater{}    \\
YIELD    & M-Coop.    & \textless{}S\textgreater{} reduced protest against \textless{}T\textgreater{}    & \textless{}S\textgreater{} promised to reduce protest for \textless{}T\textgreater{}    \\
YIELD    & M-Coop.    & \textless{}S\textgreater{} released person of \textless{}T\textgreater{}    & \textless{}S\textgreater{} promised to release person of \textless{}T\textgreater{}    \\
YIELD    & M-Coop.    & \textless{}S\textgreater{} resigned from the position in \textless{}T\textgreater{}    & \textless{}S\textgreater{} promised to resign from the position in \textless{}T\textgreater{}    \\
YIELD    & M-Coop.    & \textless{}S\textgreater{} retreated forces from \textless{}T\textgreater{}    & \textless{}S\textgreater{} promised to retreat forces from \textless{}T\textgreater{}    \\
YIELD    & M-Coop.    & \textless{}S\textgreater{} returned property of \textless{}T\textgreater{}    & \textless{}S\textgreater{} promised to return property of \textless{}T\textgreater{}    \\
YIELD    & M-Coop.    & \textless{}S\textgreater{} surrendered to \textless{}T\textgreater{}    & \textless{}S\textgreater{} promised to surrender to \textless{}T\textgreater{}    \\
YIELD    & M-Coop.    & \textless{}S\textgreater{} undertook reform in \textless{}T\textgreater{}    & \textless{}S\textgreater{} promised to undertake reform in \textless{}T\textgreater{}    \\
ACCUSE    & V-Conf.    & \textless{}S\textgreater{} accused \textless{}T\textgreater{} of something    & None    \\
ACCUSE    & V-Conf.    & \textless{}S\textgreater{} brought lawsuit against \textless{}T\textgreater{}    & None    \\
ACCUSE    & V-Conf.    & \textless{}S\textgreater{} expressed complaints of \textless{}T\textgreater{}    & None    \\
REQUEST    & V-Conf.    & \textless{}S\textgreater{} demanded something from \textless{}T\textgreater{}    & None    \\
INVESTIGATE & V-Conf.    & \textless{}S\textgreater{} investigated something of \textless{}T\textgreater{}    & \textless{}S\textgreater{} planned to investigate something of \textless{}T\textgreater{}    \\
INVESTIGATE & V-Conf.    & \textless{}S\textgreater{} sent people to investigate \textless{}T\textgreater{}    & \textless{}S\textgreater{} planned to send people to investigate \textless{}T\textgreater{}    \\
REJECT    & V-Conf.    & \textless{}S\textgreater{} defied laws of \textless{}T\textgreater{}    & None    \\
REJECT    & V-Conf.    & \textless{}S\textgreater{} rejected proposals of \textless{}T\textgreater{}    & None    \\
REJECT    & V-Conf.    & \textless{}S\textgreater{} rejected cooperation with \textless{}T\textgreater{}    & None    \\
REJECT    & V-Conf.    & \textless{}S\textgreater{} rejected to do something for \textless{}T\textgreater{}    & None    \\
REJECT    & V-Conf.    & \textless{}S\textgreater{} rejected to stop something against \textless{}T\textgreater{}    & None    \\
REJECT    & V-Conf.    & \textless{}S\textgreater{} rejected to consult with \textless{}T\textgreater{}    & None    \\
REJECT    & V-Conf.    & \textless{}S\textgreater{} rejected to yield to \textless{}T\textgreater{}    & None    \\
THREATEN    & V-Conf.    & \textless{}S\textgreater{} issued a ultimatum to \textless{}T\textgreater{}    & None    \\
THREATEN    & V-Conf.    & \textless{}S\textgreater{} threatened something against \textless{}T\textgreater{}    & None    \\
COERCE    & M-Conf.    & \textless{}S\textgreater{} arrested person of \textless{}T\textgreater{}    & \textless{}S\textgreater{} threatened to arrest person of \textless{}T\textgreater{}    \\
COERCE    & M-Conf.    & \textless{}S\textgreater{} attacked \textless{}T\textgreater{} cybernetically    & \textless{}S\textgreater{} threatened to attack \textless{}T\textgreater{} cybernetically    \\
COERCE    & M-Conf.    & \textless{}S\textgreater{} deported person of \textless{}T\textgreater{}    & \textless{}S\textgreater{} threatened to deport person of \textless{}T\textgreater{}    \\
COERCE    & M-Conf.    & \textless{}S\textgreater{} detained person of \textless{}T\textgreater{}    & \textless{}S\textgreater{} threatened to detain person of \textless{}T\textgreater{}    \\
COERCE    & M-Conf.    & \textless{}S\textgreater{} imposed blockades in \textless{}T\textgreater{}    & \textless{}S\textgreater{} threatened to impose blockades in \textless{}T\textgreater{}    \\
COERCE    & M-Conf.    & \textless{}S\textgreater{} imposed state of emergency in \textless{}T\textgreater{}    & \textless{}S\textgreater{} threatened to impose state of emergency in \textless{}T\textgreater{}    \\
COERCE    & M-Conf.    & \textless{}S\textgreater{} imposed more restrictions on \textless{}T\textgreater{}    & \textless{}S\textgreater{} threatened to impose restrictions on \textless{}T\textgreater{}    \\
COERCE    & M-Conf.    & \textless{}S\textgreater{} repressed person of \textless{}T\textgreater{}    & \textless{}S\textgreater{} threatened to repress person of \textless{}T\textgreater{}    \\
COERCE    & M-Conf.    & \textless{}S\textgreater{} seized property of \textless{}T\textgreater{}    & \textless{}S\textgreater{} threatened to seize property of \textless{}T\textgreater{}    \\
ASSAULT    & M-Conf.    & \textless{}S\textgreater{} seized territory of \textless{}T\textgreater{}    & \textless{}S\textgreater{} threatened to seize territory of \textless{}T\textgreater{}    \\
ASSAULT    & M-Conf.    & \textless{}S\textgreater{} assaulted person of \textless{}T\textgreater{}    & \textless{}S\textgreater{} threatened to assault person of \textless{}T\textgreater{}    \\
ASSAULT    & M-Conf.    & \textless{}S\textgreater{} destropyed property of \textless{}T\textgreater{}    & \textless{}S\textgreater{} threatened to destropy property of \textless{}T\textgreater{}    \\
ASSAULT    & M-Conf.    & \textless{}S\textgreater{} killed person of \textless{}T\textgreater{}    & \textless{}S\textgreater{} threatened to kill person of \textless{}T\textgreater{}    \\
ASSAULT    & M-Conf.    & \textless{}S\textgreater{} launched military strikes against \textless{}T\textgreater{}    & \textless{}S\textgreater{} threatened to launch military strikes against \textless{}T\textgreater{}    \\
ASSAULT    & M-Conf.    & \textless{}S\textgreater{} violated ceasefire with \textless{}T\textgreater{}    & \textless{}S\textgreater{} threatened to violate ceasefire with \textless{}T\textgreater{}    \\
FIGHT    & M-Conf.    & \textless{}S\textgreater{} attempted to assassinate \textless{}T\textgreater{}    & None    \\
FIGHT    & M-Conf.    & \textless{}S\textgreater{} used person of \textless{}T\textgreater{} as human shield    & None    \\
FIGHT    & M-Conf.    & Explosives in \textless{}S\textgreater{} attacked \textless{}T\textgreater{}    & None    \\
MOBILIZE    & M-Conf.    & \textless{}S\textgreater{} increased forces in \textless{}T\textgreater{}    & \textless{}S\textgreater{} threatened to increase forces in \textless{}T\textgreater{}    \\
MOBILIZE    & M-Conf.    & \textless{}S\textgreater{} kept alert in \textless{}T\textgreater{}    & \textless{}S\textgreater{} threatened to keep alert in \textless{}T\textgreater{}    \\
MOBILIZE    & M-Conf.    & \textless{}S\textgreater{} prepared forces against \textless{}T\textgreater{}    & \textless{}S\textgreater{} threatened to prepare forces against \textless{}T\textgreater{}    \\
PROTEST    & M-Conf.    & \textless{}S\textgreater{} launched protests against \textless{}T\textgreater{}    & \textless{}S\textgreater{} threatened to launch protests against \textless{}T\textgreater{}    \\
PROTEST    & M-Conf.    & \textless{}S\textgreater{} launched protests in \textless{}T\textgreater{}    & \textless{}S\textgreater{} threatened to launch protests in \textless{}T\textgreater{}    \\
PROTEST    & M-Conf.    & \textless{}S\textgreater{} protestors obstructed roads against \textless{}T\textgreater{}    & \textless{}S\textgreater{} protestors threatened to obstruct roads against \textless{}T\textgreater{}    \\
PROTEST    & M-Conf.    & \textless{}S\textgreater{} undertook boycotts against \textless{}T\textgreater{}    & \textless{}S\textgreater{} threatened to undertake boycott against \textless{}T\textgreater{}    \\
SANCTION    & M-Conf.    & \textless{}S\textgreater{} discontinued cooperation with \textless{}T\textgreater{}    & \textless{}S\textgreater{} threatened to discontinue cooperation with \textless{}T\textgreater{}    \\
SANCTION    & M-Conf.    & \textless{}S\textgreater{} expelled diplomatic people of \textless{}T\textgreater{}    & \textless{}S\textgreater{} threatened to expel diplomatic people of \textless{}T\textgreater{}    \\
SANCTION    & M-Conf.    & \textless{}S\textgreater{} expelled organizations of \textless{}T\textgreater{}    & \textless{}S\textgreater{} threatened to expel organizations of \textless{}T\textgreater{}    \\
SANCTION    & M-Conf.    & \textless{}S\textgreater{} expelled peacekeepers of \textless{}T\textgreater{}    & \textless{}S\textgreater{} threatened to expel peacekeepers of \textless{}T\textgreater{}    \\
SANCTION    & M-Conf.    & \textless{}S\textgreater{} halted negotiations with \textless{}T\textgreater{}    & \textless{}S\textgreater{} threatened to halt negotiate with \textless{}T\textgreater{}    \\
SANCTION    & M-Conf.    & \textless{}S\textgreater{} reduced aid to \textless{}T\textgreater{}    & \textless{}S\textgreater{} threatened to reduce aid to \textless{}T\textgreater{}    \\
SANCTION    & M-Conf.    & \textless{}S\textgreater{} retreated peace forces from \textless{}T\textgreater{}    & \textless{}S\textgreater{} threatened to retreat peace forces from \textless{}T\textgreater{}    \\   
\bottomrule
\end{tabular}
}
 \caption{The mode-aware hypothesis table considering Past and Future modes. <S> and <T> represent the source and the  target entities in practical examples. Some hypotheses do not require Future variants as their labels (Rootcode and Quadcode) remain unchanged from Past to Future, as indicated in Table \ref{tab:plover_mode}. }\label{tab:mode-aware hypothese table}
\end{table*}

\begin{table*}[t]
\centering
\small
\begin{tabular}{p{16cm}}

\toprule

\textbf{Relation Extraction (RE)} Task is to classify the political relations between a source (indicated by  \textless{}S\textgreater{}\textless{}/S\textgreater{}) and a target (indicated by  \textless{}T\textgreater{}\textless{}/T\textgreater{}) within a given input sentence. 
The goal is to assign these relations into a predefined set of labels. The predefined set of relation labels 1-15 is as follows. The relations can be categorized into four quadrants: Q1(Verbal Cooperation), Q2 (Material Cooperation), Q3 (Verbal Conflict), and Q4 (Material Conflict). 

\\
1. AGREE, Q1: Agree to, offer, promise, or otherwise indicate willingness or commitment to cooperate, including promises to sign or ratify agreements. Cooperative actions (CONSULT, SUPPORT, COOPERATE, AID, YIELD) reported in future tense are also taken to imply intentions and should be coded as AGREE.  
\\
2. CONSULT, Q1: All consultations and meetings, including visiting and hosting visits, meeting at neutral location, and consultation by phone or other media. 
\\
3. SUPPORT, Q1: Initiate, resume, improve, or expand diplomatic, non-material cooperation; express support for, commend, approve policy, action, or actor, or ratify, sign, or finalize an agreement or treaty.
\\
4. COOPERATE, Q2: Initiate, resume, improve, or expand mutual material cooperation or exchange, including economics, military, judicial matters, and sharing of intelligence.
\\
5. AID, Q2: All provisions of providing material aid whose material benefits primarily accrue to the recipient, including monetary, military,humanitarian, asylum etc.
\\
6. YIELD, Q2: yieldings or concessions, such as resignations of government officials, easing of legal restrictions, the release of prisoners, repatriation of refugees or property, allowing third party access, disarming militarily, implementing a ceasefire, and a military retreat.
\\
7. REQUEST, Q3: All verbal requests, demands, and orders, which are less forceful than threats and potentially carry less serious repercussions. Demands that take the form of demonstrations, protests, etc. are coded as PROTEST.
\\
8. ACCUSE, Q3: Express disapprovals, objections, and complaints; condemn, decry a policy or an action; criticize, defame, denigrate responsible parties. Accuse, allege, or charge, both judicially and informally. Sue or bring to court. Investigations.
\\
9. REJECT, Q3: All rejections and refusals, such as assistance, changes in policy, yielding, or meetings.
\\
10. THREATEN, Q3: All threats, coercive or forceful warnings with serious potential repercussions. Threats are generally verbal acts except for purely symbolic material actions such as having an unarmed group place a flag on some territory.
\\
11. PROTEST, Q4: All civilian demonstrations and other collective actions carried out as protests against the recipient: Dissent collectively, publicly show negative feelings or opinions; rally, gather to protest a policy, action, or actor(s).
\\
12. SANCTION, Q4: All reductions in existing, routine, or cooperative relations. For example, withdrawing or discontinuing diplomatic, commercial, or material exchanges.
\\
13. MOBILIZE, Q4: All military or police moves that fall short of the actual use of force. This category is different from ASSAULT, which refers to actual uses of force, while military posturing falls short of actual use of force and is typically a demonstration of military capabilities and readiness. MOBILIZE is also distinct from THREAT in that the latter is typically verbal, and does not involve any activity that is undertaken to demonstrate military power.
\\
14. COERCE, Q4: Repression, restrictions on rights, or coercive uses of power falling short of violence, such as arresting, deporting, banning individuals, imposing curfew, imposing restrictions on political freedoms or movement, conducting cyber attacks, etc.
\\
15. ASSAULT, Q4: Deliberate actions which can potentially result in substantial physical harm. 
\\
\\
Note that we give priority to labels in Material Conflict over Verbal Conflict.
For example,  we label ``protest to request'' as material PROTEST other than verbal REQUEST.
Similarly, we label ``convict and arrest'' as material COERCE other than verbal ACCUSE, considering the more severe actions involved.
\\
\midrule
\textbf{Input and Task Requirement: }

Perform the RE task for the given input list and print the output with columns (No., Label, Quadrants) split by the tab delimiter. Use 1-15 to denote the predefined labels above (1. AGREE, 2. CONSULT, 3. SUPPORT, 4. COOPERATE, 5. AID, 6. YIELD,  7. REQUEST,  8. ACCUSE, 9. REJECT,  10. THREATEN, 11. PROTEST, 12. SANCTION,  13. MOBILIZE, 14. COERCE, and 15. ASSAULT).\\
\end{tabular}

\begin{tabular}{cp{15cm}}
\textbf{No.} & \textbf{Sentence} \\
1 &	\textless{}S\textgreater{}A Brazilian federal court\textless{}/S\textgreater{} has rejected a request from \textless{}T\textgreater{}jailed former President Luiz Inacio Lula da Silva\textless{}/T\textgreater{} to be present at the first debate of presidential candidates for October's election.\\
2 & \textless{}S\textgreater{}Afghan rebels\textless{}/S\textgreater{} have kidnapped up to 16 \textless{}T\textgreater{}Soviet civilian advisers\textless{}/T\textgreater{} from a town bazaar and exploded a series of bombs in the capital Kabul, western diplomatic sources in neighboring Pakistan said today.\\
3 &	\textless{}S\textgreater{}A local Taliban leader and his five associates\textless{}/S\textgreater{} have given up fighting and surrendered in \textless{}T\textgreater{}Afghanistan’s northern Faryab province\textless{}/T\textgreater{}, an army source said Tuesday.\\

4 & \textless{}S\textgreater{}French National Assembly president Laurent Fabius\textless{}/S\textgreater{} and a group of deputies held talks with leaders of\textless{}T\textgreater{}Romania's\textless{}/T\textgreater{} new government on Tuesday, the first high level Western delegation to visit Bucharest since last month's revolution.
\end{tabular}

\begin{tabular}{p{16cm}}
\midrule
\textbf{Output:}
\end{tabular}

\begin{tabular}{cllc}
\textbf{No.} & \textbf{Label} & \textbf{Quadrants}  & \textbf{Correct?}\\
1 & 9 (REJECT) &  Q3: Verbal Conflict  &  \checkmark\\
2 & 15 (ASSAULT) & Q4: Material Conflict  &  \checkmark\\
3 & 6 (YIELD) &   Q2: Material Cooperation  &  \checkmark\\
4 & 2 (CONSULT)    &  Q1: Verbal Cooperation  &  \checkmark 
\end{tabular}

\begin{tabular}{p{16cm}}
\bottomrule\\
\end{tabular}

\caption{Input and Output of ChatGPT.}
\label{tab:chatgpt}
\end{table*}

\end{document}